\documentclass[]{book}
\usepackage{lmodern}
\usepackage{amssymb,amsmath}
\usepackage{ifxetex,ifluatex}
\usepackage{fixltx2e} 
\ifnum 0\ifxetex 1\fi\ifluatex 1\fi=0 
  \usepackage[T1]{fontenc}
  \usepackage[utf8]{inputenc}
\else 
  \ifxetex
    \usepackage{mathspec}
  \else
    \usepackage{fontspec}
  \fi
  \defaultfontfeatures{Ligatures=TeX,Scale=MatchLowercase}
\fi
\IfFileExists{upquote.sty}{\usepackage{upquote}}{}
\IfFileExists{microtype.sty}{%
\usepackage{microtype}
\UseMicrotypeSet[protrusion]{basicmath} 
}{}
\usepackage[margin=1in]{geometry}
\usepackage{hyperref}
\PassOptionsToPackage{usenames,dvipsnames}{color} 
\hypersetup{unicode=true,
            pdftitle={Corpus Phonetics Tutorial},
            pdfauthor={Eleanor Chodroff},
            colorlinks=true,
            linkcolor=Maroon,
            citecolor=Blue,
            urlcolor=blue,
            breaklinks=true}
\usepackage{natbib}
\usepackage{color}
\usepackage{fancyvrb}

\DefineVerbatimEnvironment{Highlighting}{Verbatim}{commandchars=\\\{\}}
\usepackage{framed}
\definecolor{shadecolor}{RGB}{248,248,248}
\newenvironment{Shaded}{\begin{snugshade}}{\end{snugshade}}

\newcommand{\DecValTok}[1]{\textcolor[rgb]{0.00,0.00,0.81}{#1}}

\newcommand{\FloatTok}[1]{\textcolor[rgb]{0.00,0.00,0.81}{#1}}

\newcommand{\CharTok}[1]{\textcolor[rgb]{0.31,0.60,0.02}{#1}}

\newcommand{\StringTok}[1]{\textcolor[rgb]{0.31,0.60,0.02}{#1}}

\newcommand{\CommentTok}[1]{\textcolor[rgb]{0.56,0.35,0.01}{\textit{#1}}}

\newcommand{\ControlFlowTok}[1]{\textcolor[rgb]{0.13,0.29,0.53}{\textbf{#1}}}
\newcommand{\OperatorTok}[1]{\textcolor[rgb]{0.81,0.36,0.00}{\textbf{#1}}}

\newcommand{\ErrorTok}[1]{\textcolor[rgb]{0.64,0.00,0.00}{\textbf{#1}}}
\newcommand{\NormalTok}[1]{#1}
\usepackage{longtable,booktabs}
\usepackage{graphicx,grffile}
\makeatletter
\def\maxwidth{\ifdim\Gin@nat@width>\linewidth\linewidth\else\Gin@nat@width\fi}
\def\maxheight{\ifdim\Gin@nat@height>\textheight\textheight\else\Gin@nat@height\fi}
\makeatother
\setkeys{Gin}{width=\maxwidth,height=\maxheight,keepaspectratio}
\IfFileExists{parskip.sty}{%
\usepackage{parskip}
}{
\setlength{\parindent}{0pt}
\setlength{\parskip}{6pt plus 2pt minus 1pt}
}
\providecommand{\tightlist}{%
  \setlength{\itemsep}{0pt}\setlength{\parskip}{0pt}}
\ifx\paragraph\undefined\else
\let\oldparagraph\paragraph
\renewcommand{\paragraph}[1]{\oldparagraph{#1}\mbox{}}
\fi
\ifx\subparagraph\undefined\else
\let\oldsubparagraph\subparagraph
\renewcommand{\subparagraph}[1]{\oldsubparagraph{#1}\mbox{}}
\fi

\let\rmarkdownfootnote\footnote%
\def\footnote{\protect\rmarkdownfootnote}

\usepackage{titling}

  \title{Corpus Phonetics Tutorial}
    \pretitle{\vspace{\droptitle}\centering\huge}
  \posttitle{\par}
    \author{Eleanor Chodroff}
    \preauthor{\centering\large\emph}
  \postauthor{\par}
      \predate{\centering\large\emph}
  \postdate{\par}
    \date{Core content written: 2015 \textbar{} Updated: 2018-11-13}

\usepackage{amsthm}

\theoremstyle{definition}

\theoremstyle{definition}

\theoremstyle{definition}

\theoremstyle{remark}

\begin{document}
\maketitle

{
\hypersetup{linkcolor=black}
\setcounter{tocdepth}{1}
\tableofcontents
}
\chapter{Introduction}\label{introduction}

Corpus phonetics has become an increasingly popular method of research in linguistic analysis. With advances in speech technology and computational power, large scale processing of speech data has become a viable technique. A fair number of researchers have exploited these methods, yet these techniques still remain elusive for many. In the words of Mark Liberman, there has been ``surprisingly little change in style and scale of {[}phonetic{]} research'' from 1966 on, implying that the field still relies on small sample sizes of speech data (2009). While ``big data'' phonetics is not the be-all and end-all of phonetic research, larger sample sizes ensure more statistically sound conclusions about phonetic values in an individual or population. Furthermore, corpus research is not synonymous with big data. Rather, corpus phonetics describes a method of processing speech data with advantages primarily gained in its computational power (relation to big data) and efficiency. The methods and tools developed for corpus phonetics are based on engineering algorithms primarily from automatic speech recognition (ASR), as well as simple programming for data manipulation. This tutorial aims to bring some of these tools to the non-engineer, and specifically to the speech scientist.

Acoustic analysis programs such as \href{http://www.fon.hum.uva.nl/praat/}{Praat}, \href{http://www.mathworks.com/products/matlab/}{MATLAB}, and \href{https://www.r-project.org/about.html}{R} (check out the tuneR and multitaper packages) are already capable of large scale phonetic measurement via their respective scripting languages. While the tutorial covers some phonetic processing in Praat, the primary aim is to introduce supplementary tools to phonetic processing. These tools are based on concepts and algorithms from automatic speech recognition, which allow for automatic alignment of phonetic boundaries to the speech signal.

In particular, the tutorial currently covers various tools from the Kaldi Automatic Speech Recognition Toolki, FAVE-align, the Montreal Forced Aligner, Penn Phonetics Lab Forced Aligner, and AutoVOT. (The documentation for the Penn Forced Aligner is marked as ``legacy'' as this system has essentially been replaced by FAVE-align.) Kaldi is an automatic speech recognition toolkit that provides the infrastructure to build personalized \textbf{acoustic models} and \textbf{forced alignment} systems. Acoustic models are the statistical representations of each phoneme's acoustic information. The ``personalized'' component means that this system is capable of modeling any corpus of speech, be it British English, Southern American English, Hungarian, or Korean. It additionally houses many speech processing algorithms, which may be of use to the speech scientist. This tutorial will cover acoustic model training and forced alignment in Kaldi; however, the toolkit as a whole provides exceptional potential for phonetic research. ``Forced alignment'' is the automatic synchronization of a sequence of phones with an audio file. This process employs \textbf{acoustic models} of the sounds of a language, along with a pronunciation lexicon which provides a canonical mapping from orthographic words to sequences of phones. Forced alignment greatly expedites data processing and phonetic measurement. Kaldi, FAVE-align, and the Montreal Forced Aligner are all capable of forced alignment, but with varying degrees of flexibility with respect to the input speech. Finally, AutoVOT is an automatic voice onset time (VOT) measurement tool that demarcates the burst release and vocalic onset of word-initial, prevocalic stop consonants.

Finally, the tutorial assumes basic familiarity with \href{http://www.fon.hum.uva.nl/praat/}{Praat}, as well as a Mac operating system, primarily for the default bash/Unix shell in the Terminal application. If using a PC, I recommend downloading \href{https://www.cygwin.com/}{Cygwin} for running bash/Unix commands. For AutoVOT and the Penn Forced Aligner, most of the Unix commands are provided in the tutorial itself. While I try to provide as many of the commands as possible, Kaldi requires more fluency in shell scripting. If you have not used the Terminal application before, I recommend looking over some basic Unix commands online (Google is every programmer's best friend). For a list of the most useful commands, I recommend this \href{http://www.tutorialspoint.com/unix/unix-useful-commands.htm}{website}. For more details regarding the argument structure, I recommend this \href{https://kb.iu.edu/d/afsk}{website}.

Each section covers the prerequisites for each program's installation, as well as a standard recipe for each program. As a good rule of thumb, all prerequisites should be installed prior to installation of the desired program.

Citations for each of the programs can be found below:

\begin{itemize}
\tightlist
\item
  Kaldi
\end{itemize}

Povey, D., Ghoshal, A., Boulianne, G., Burget, L., Glembek, O., Goel, N., Hannemann, M., Motlicek, P., Qian, Y., Schwartz, P., Silovsky, J., Stemmer, G., \& Vesely, K. (2011). The Kaldi speech recognition toolkit. In IEEE 2011 Workshop on ASRU.

\begin{Shaded}
\begin{Highlighting}[]
\OperatorTok{@}\NormalTok{INPROCEEDINGS\{}
\NormalTok{         Povey_ASRU2011,}
\NormalTok{         author =}\StringTok{ }\NormalTok{\{Povey, Daniel and Ghoshal, Arnab and Boulianne, Gilles and }
\NormalTok{           Burget, Lukas and Glembek, Ondrej and Goel, Nagendra and }
\NormalTok{           Hannemann, Mirko and Motlicek, Petr and Qian, Yanmin and }
\NormalTok{           Schwarz, Petr and Silovsky, Jan and Stemmer, Georg and Vesely, Karel\},}
\NormalTok{       keywords =}\StringTok{ }\NormalTok{\{ASR, Automatic Speech Recognition, GMM, HTK, SGMM\},}
\NormalTok{          month =}\StringTok{ }\NormalTok{dec,}
\NormalTok{          title =}\StringTok{ }\NormalTok{\{The Kaldi Speech Recognition Toolkit\},}
\NormalTok{      booktitle =}\StringTok{ }\NormalTok{\{IEEE }\DecValTok{2011}\NormalTok{ Workshop on Automatic Speech Recognition and Understanding\},}
\NormalTok{           year =}\StringTok{ }\NormalTok{\{}\DecValTok{2011}\NormalTok{\},}
\NormalTok{      publisher =}\StringTok{ }\NormalTok{\{IEEE Signal Processing Society\},}
\NormalTok{       location =}\StringTok{ }\NormalTok{\{Hilton Waikoloa Village, Big Island, Hawaii, US\},}
\NormalTok{           note =}\StringTok{ }\NormalTok{\{IEEE Catalog No.}\OperatorTok{:}\StringTok{ }\NormalTok{CFP11SRW}\OperatorTok{-}\NormalTok{USB\},}
\NormalTok{\}}
\end{Highlighting}
\end{Shaded}

\begin{itemize}
\tightlist
\item
  FAVE-align
\end{itemize}

Rosenfelder, Ingrid; Fruehwald, Josef; Evanini, Keelan; Seyfarth, Scott; Gorman, Kyle; Prichard, Hilary; Yuan, Jiahong; 2014. FAVE (Forced Alignment and Vowel Extraction) Program Suite v1.2.2 10.5281/zenodo.22281

\begin{itemize}
\tightlist
\item
  Montreal Forced Aligner
\end{itemize}

McAuliffe, Michael, Michaela Socolof, Sarah Mihuc, Michael Wagner, and Morgan Sonderegger (2017). Montreal Forced Aligner {[}Computer program{]}. Version 0.9.0, retrieved 17 January 2017 from \url{http://montrealcorpustools.github.io/Montreal-Forced-Aligner/}.

\begin{itemize}
\tightlist
\item
  Penn Phonetics Lab Forced Aligner
\end{itemize}

Yuan, Jiahong., \& Liberman, Mark. (2008). Speaker identification on the SCOTUS corpus. In Proceedings of Acoustics, '08.

\begin{itemize}
\tightlist
\item
  AutoVOT
\end{itemize}

Keshet, J., Sonderegger, M., Knowles, T. (2014). AutoVOT: A tool for automatic measurement of voice onset time using discriminative structured prediction {[}Computer program{]}. Version 0.91, retrieved August 2014 from \url{https://github.com/mlml/autovot/}.

\chapter{Kaldi}\label{kaldi}

\section{Overview}\label{overview}

\textbf{What is Kaldi?} Kaldi is a state-of-the-art automatic speech recognition (ASR) toolkit, containing almost any algorithm currently used in ASR systems. It also contains recipes for training your own acoustic models on commonly used speech corpora such as the Wall Street Journal Corpus, TIMIT, and more. These recipes can also serve as a template for training acoustic models on your own speech data.

\textbf{What are acoustic models?} Acoustic models are the statistical representations of a phoneme's acoustic information. A phoneme here represents a member of the set of speech sounds in a language. N.B., this use of the term `phoneme' only loosely corresponds to the linguistic use of the term `phoneme'.

The acoustic models are created by training the models on acoustic features from labeled data, such as the Wall Street Journal Corpus, TIMIT, or any other transcribed speech corpus. There are many ways these can be trained, and the tutorial will try to cover some of the more standard methods. Acoustic models are necessary not only for automatic speech recognition, but also for forced alignment.

Kaldi provides tremendous flexibility and power in training your own acoustic models and forced alignment system. The following tutorial covers a general recipe for training on your own data. This part of the tutorial assumes more familiarity with the terminal; you will also be much better off if you can program basic text manipulations.

Please also refer to the \href{http://www.kaldi-asr.org/doc/}{Kaldi website} for thorough documentation.

\section{Installation}\label{installation}

Please refer to \url{http://www.kaldi-asr.org/doc/install.html} for more details.

\begin{enumerate}
\def\labelenumi{\arabic{enumi}.}
\tightlist
\item
  Prerequisites
\end{enumerate}

\begin{itemize}
\tightlist
\item
  Git
\end{itemize}

Git is a version control system that let's developers update source code and easily redistribute updates to the users. Git can be installed via homebrew or following instructions \href{https://git-scm.com/book/en/v1/Getting-Started-Installing-Git}{here}.

\begin{itemize}
\tightlist
\item
  Subversion (svn)
\end{itemize}

Subversion is also a \href{https://en.wikipedia.org/wiki/Revision_control}{version control system} that keeps track of individual changes while developing the source code. Some of the example scripts still depend on this package.

\begin{enumerate}
\def\labelenumi{\arabic{enumi}.}
\setcounter{enumi}{1}
\tightlist
\item
  Downloading
\end{enumerate}

It is recommended that Kaldi be installed on a machine with good computing power. Following the instructions for downloading Kaldi on this page: \url{http://kaldi-asr.org/doc/install.html}, first direct the terminal to where you would like to install Kaldi, and then type the following:

\begin{Shaded}
\begin{Highlighting}[]
\NormalTok{git clone https}\OperatorTok{:}\ErrorTok{//}\NormalTok{github.com}\OperatorTok{/}\NormalTok{kaldi}\OperatorTok{-}\NormalTok{asr}\OperatorTok{/}\NormalTok{kaldi.git kaldi }\OperatorTok{--}\NormalTok{origin upstream}
\end{Highlighting}
\end{Shaded}

\begin{enumerate}
\def\labelenumi{\arabic{enumi}.}
\setcounter{enumi}{2}
\tightlist
\item
  Installation
\end{enumerate}

Locate the file \texttt{INSTALL} in the downloaded package and follow the instructions there. In short, you'll need to follow the install instructions in \texttt{kaldi/tools} and then in \texttt{kaldi/src}. The most typical installation should involve the following code, but read the \texttt{INSTALL} file just in case:

\begin{Shaded}
\begin{Highlighting}[]
\NormalTok{cd kaldi}\OperatorTok{/}\NormalTok{tools  }
\NormalTok{extras}\OperatorTok{/}\NormalTok{check_dependencies.sh  }
\NormalTok{make}

\NormalTok{cd kaldi}\OperatorTok{/}\NormalTok{src  }
\NormalTok{.}\OperatorTok{/}\NormalTok{configure  }
\NormalTok{make depend  }
\NormalTok{make}
\end{Highlighting}
\end{Shaded}

\section{Familiarization}\label{familiarization}

This section serves as a cursory overview of Kaldi's directory structure. The top-level directories are \texttt{egs}, \texttt{src}, \texttt{tools}, \texttt{misc}, and \texttt{windows}. The directories we will be using are \texttt{egs} and \texttt{src}.

\texttt{egs} stands for `examples' and contains example training recipes for most major speech corpora. Training recipes are available for the Wall Street Journal Corpus (\texttt{wsj}), TIMIT (\texttt{timit}), Resource Management (\texttt{rm}), and many others. Under each of these directories are usually a few different versions (\texttt{s3}, \texttt{s4}, \texttt{s5}, etc.) The highest number, usually \texttt{s5}, is the most current version and should be used for any new development or training. The older versions are kept for archival purposes only.

\texttt{src} stands for `source' or `source code' and contains most of the source code for programs that the training recipes call.

For each training recipe directory, there is a standard sub-directory structure. This is best exemplified in the Resource Management directory (\texttt{egs/rm/s5}). The top directory contains the run script (\texttt{run.sh}), as well as two other required scripts (\texttt{cmd.sh} and \texttt{path.sh}). The sub-directories are \texttt{conf} (configuration), \texttt{data}, \texttt{exp} (experiments), \texttt{local}, \texttt{steps}, and \texttt{utils} (utilities). The directories we will primarily be using are \texttt{data} and \texttt{exp}. The \texttt{data} directory will eventually house information relevant to your own data such as transcripts, dictionaries, etc. The \texttt{exp} directory will eventually contain the output of the training and alignment scripts, or the acoustic models.

\begin{figure}
\centering
\includegraphics{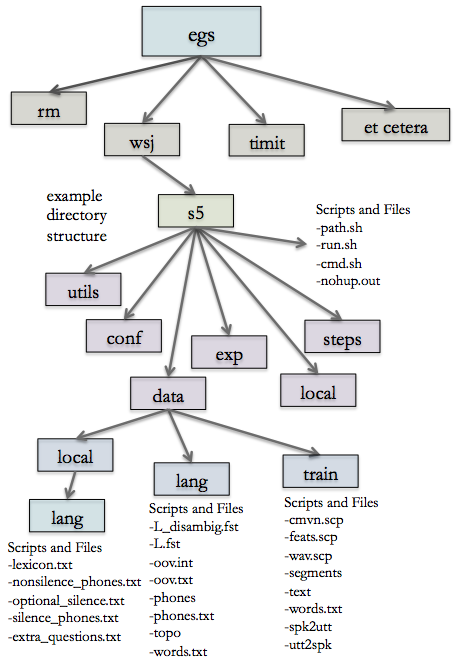}
\caption{Example directory structure}
\end{figure}

\section{Training Overview}\label{training-overview}

Before diving into the scripts, it is essential to understand the basic procedure for training acoustic models. Given the audience and purpose of the tutorial, this section will focus on the process as opposed to the computation (see \href{http://www.amazon.com/Speech-Language-Processing-Daniel-Jurafsky/dp/0131873210/ref=sr_1_1?s=books\&ie=UTF8\&qid=1435870892\&sr=1-1\&keywords=speech+and+language+processing\&pebp=1435870888175\&perid=0HS6VNBZX7NEZN1NTCX2}{Jurafsky and Martin 2008}, \href{https://ieeexplore.ieee.org/abstract/document/536824/}{Young 1996}, among many others). The procedure can be laid out as follows:

\textbf{1. Obtain a written transcript of the speech data}

For a more precise alignment, utterance (\textasciitilde{}sentence) level start and end times are helpful, but not necessary.

\textbf{2. Format transcripts for Kaldi}

Kaldi requires various formats of the transcripts for acoustic model training. You'll need the start and end times of each utterance, the speaker ID of each utterance, and a list of all words and phonemes present in the transcript.

\textbf{3. Extract acoustic features from the audio}

Mel Frequency Cepstral Coefficients (MFCC) are the most commonly used features, but Perceptual Linear Prediction (PLP) features and other features are also an option. These features serve as the basis for the acoustic models.

\textbf{4. Train monophone models}

A monophone model is an acoustic model that does not include any contextual information about the preceding or following phone. It is used as a building block for the triphone models, which do make use of contextual information.

*Note: from this point forward, we will be assuming a Gaussian Mixture Model/Hidden Markov Model (GMM/HMM) framework. This is in contrast to a deep neural network (DNN) system.

\textbf{5. Align audio with the acoustic models}

The parameters of the acoustic model are estimated in acoustic training steps; however, the process can be better optimized by cycling through training and alignment phases. This is also known as Viterbi training (related, but more computationally expensive procedures include the Forward-Backward algorithm and Expectation Maximization). By aligning the audio to the reference transcript with the most current acoustic model, additional training algorithms can then use this output to improve or refine the parameters of the model. Therefore, each training step will be followed by an alignment step where the audio and text can be realigned.

\textbf{6. Train triphone models}

While monophone models simply represent the acoustic parameters of a single phoneme, we know that phonemes will vary considerably depending on their particular context. The triphone models represent a phoneme variant in the context of two other (left and right) phonemes.

At this point, we'll also need to deal with the fact that not all triphone units are present (or will ever be present) in the dataset. There are (\# of phonemes)\textsuperscript{3} possible triphone models, but only a subset of those will actually occur in the data. Furthermore, the unit must also occur multiple times in the data to gather sufficient statistics for the data. A phonetic decision tree groups these triphones into a smaller amount of acoustically distinct units, thereby reducing the number of parameters and making the problem computationally feasible.

\textbf{7. Re-align audio with the acoustic models \& re-train triphone models}

Repeat steps 5 and 6 with additional triphone training algorithms for more refined models. These typically include delta+delta-delta training, LDA-MLLT, and SAT. The alignment algorithms include speaker independent alignments and FMLLR.

\begin{itemize}
\tightlist
\item
  \textbf{Training Algorithms}
\end{itemize}

\textbf{Delta+delta-delta training} computes delta and double-delta features, or dynamic coefficients, to supplement the MFCC features. Delta and delta-delta features are numerical estimates of the first and second order derivatives of the signal (features). As such, the computation is usually performed on a larger window of feature vectors. While a window of two feature vectors would probably work, it would be a very crude approximation (similar to how a delta-difference is a very crude approximation of the derivative). Delta features are computed on the window of the original features; the delta-delta are then computed on the window of the delta-features.

\textbf{LDA-MLLT} stands for Linear Discriminant Analysis -- Maximum Likelihood Linear Transform. The Linear Discriminant Analysis takes the feature vectors and builds HMM states, but with a reduced feature space for all data. The Maximum Likelihood Linear Transform takes the reduced feature space from the LDA and derives a unique transformation for each speaker. MLLT is therefore a step towards speaker normalization, as it minimizes differences among speakers.

\textbf{SAT} stands for Speaker Adaptive Training. SAT also performs speaker and noise normalization by adapting to each specific speaker with a particular data transform. This results in more homogenous or standardized data, allowing the model to use its parameters on estimating variance due to the phoneme, as opposed to the speaker or recording environment.

\begin{itemize}
\tightlist
\item
  \textbf{Alignment Algorithms}
\end{itemize}

The actual alignment algorithm will always be the same; the different scripts accept different types of acoustic model input.

Speaker independent alignment, as it sounds, will exclude speaker-specific information in the alignment process.

\textbf{fMLLR} stands for Feature Space Maximum Likelihood Linear Regression. After SAT training, the acoustic model is no longer trained on the original features, but on speaker-normalized features. For alignment, we essentially have to remove the speaker identity from the features by estimating the speaker identity (with the inverse of the fMLLR matrix), then removing it from the model (by multiplying the inverse matrix with the feature vector). These quasi-speaker-independent acoustic models can then be used in the alignment process.

\section{Training Acoustic Models}\label{training-acoustic-models}

\subsection{Prepare directories}\label{prepare-directories}

Create a directory to house your training data and models:

\begin{Shaded}
\begin{Highlighting}[]
\NormalTok{cd kaldi}\OperatorTok{/}\NormalTok{egs}
\NormalTok{mkdir mycorpus}
\end{Highlighting}
\end{Shaded}

The goal of the next few sections is to recreate the directory structure laid out in Section \ref{familiarization} on Familiarization. The structure we'll be building in this section starts at the node \texttt{mycorpus}:

\begin{figure}
\centering
\includegraphics{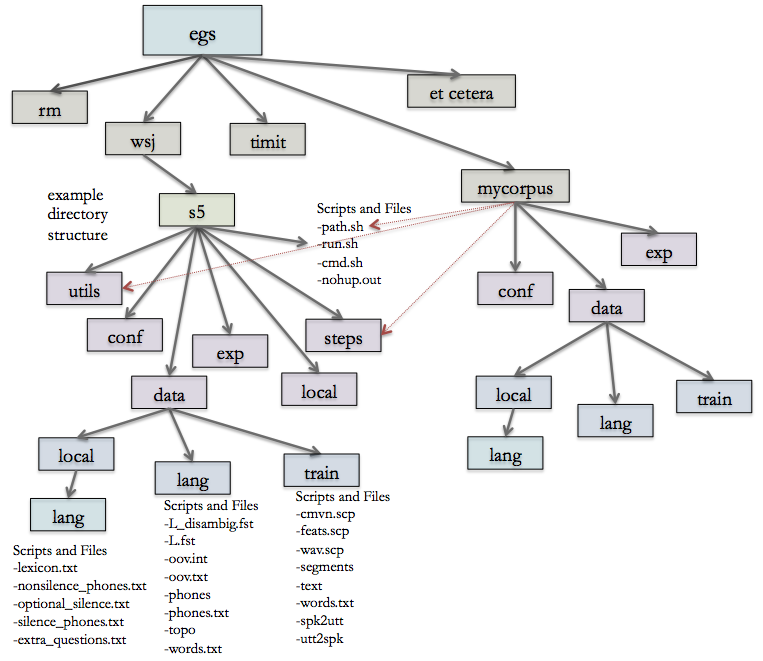}
\caption{Directory structure to replicate}
\end{figure}

In the following sections, we'll fill these directories in. For now, let's just create them.

Enter your new directory and make soft links to the following directories in the \texttt{wsj} directory to access necessary scripts: \texttt{steps}, \texttt{utils}, and \texttt{src}. In addition to the directories, you will also need a copy of the \texttt{path.sh} script in your \texttt{mycorpus} directory. \textbf{You will likely need to edit \texttt{path.sh} to make sure the KALDI-ROOT path is correct}. Make sure that the number of double dot levels takes you from your primary Kaldi directory (KALDI-ROOT) down to your working directory. For example, there are three levels between \texttt{kaldi} and \texttt{wsj/s5}, but only two levels between \texttt{kaldi} and \texttt{mycorpus}.

\begin{Shaded}
\begin{Highlighting}[]
\NormalTok{cd mycorpus}
\NormalTok{ln }\OperatorTok{-}\NormalTok{s ..}\OperatorTok{/}\NormalTok{wsj}\OperatorTok{/}\NormalTok{s5}\OperatorTok{/}\NormalTok{steps .}
\NormalTok{ln }\OperatorTok{-}\NormalTok{s ..}\OperatorTok{/}\NormalTok{wsj}\OperatorTok{/}\NormalTok{s5}\OperatorTok{/}\NormalTok{utils .}
\NormalTok{ln }\OperatorTok{-}\NormalTok{s ..}\OperatorTok{/}\NormalTok{..}\OperatorTok{/}\NormalTok{src .}
                    
\NormalTok{cp ..}\OperatorTok{/}\NormalTok{wsj}\OperatorTok{/}\NormalTok{s5}\OperatorTok{/}\NormalTok{path.sh .}
\end{Highlighting}
\end{Shaded}

Since the mycorpus directory is a level higher than \texttt{wsj/s5}, we need to edit the \texttt{path.sh} file.

\begin{Shaded}
\begin{Highlighting}[]
\NormalTok{vim path.sh}

\CommentTok{# Press i to insert; esc to exit insert mode; }
\CommentTok{# ‘:wq’ to write and quit; ‘:q’ to quit normally; }
\CommentTok{# ‘:q!’ to quit forcibly (without saving)}

\CommentTok{# Change the path line in path.sh from:}
\NormalTok{export KALDI_ROOT=}\StringTok{'pwd'}\OperatorTok{/}\NormalTok{..}\OperatorTok{/}\NormalTok{..}\OperatorTok{/}\NormalTok{.. }
\CommentTok{# to:}
\NormalTok{export KALDI_ROOT=}\StringTok{'pwd'}\OperatorTok{/}\NormalTok{..}\OperatorTok{/}\NormalTok{..}
\end{Highlighting}
\end{Shaded}

Finally, you will need to create the following directories in \texttt{mycorpus}: \texttt{exp}, \texttt{conf}, \texttt{data}. Within \texttt{data}, create the following directories: \texttt{train}, \texttt{lang}, \texttt{local} and \texttt{local/lang}. The next few steps in the tutorial will explain how to fill these directories in.

\begin{Shaded}
\begin{Highlighting}[]
\NormalTok{cd mycorpus}
\NormalTok{mkdir exp}
\NormalTok{mkdir conf}
\NormalTok{mkdir data}
                    
\NormalTok{cd data}
\NormalTok{mkdir train}
\NormalTok{mkdir lang}
\NormalTok{mkdir local}
                    
\NormalTok{cd local}
\NormalTok{mkdir lang}
\end{Highlighting}
\end{Shaded}

\subsection{\texorpdfstring{Create files for \texttt{data/train}}{Create files for data/train}}\label{create-files-for-datatrain}

The files in \texttt{data/train} contain information regarding the specifics of the audio files, transcripts, and speakers. Specifically, it will contain the following files:

\begin{itemize}
\tightlist
\item
  \texttt{text}
\item
  \texttt{segments}
\item
  \texttt{wav.scp}
\item
  \texttt{utt2spk}
\item
  \texttt{spk2utt}
\end{itemize}

\subsubsection{text}\label{text}

The \texttt{text} file is essentially the utterance-by-utterance transcript of the corpus. This is a text file with the following format:

\hypertarget{textfile}{}
utt\_id WORD1 WORD2 WORD3 WORD4 \ldots{}

utt\_id = utterance ID

Example text file:

\hypertarget{textfile}{}
110236\_20091006\_82330\_F\_0001 I'M WORRIED ABOUT THAT\\
110236\_20091006\_82330\_F\_0002 AT LEAST NOW WE HAVE THE BENEFIT\\
110236\_20091006\_82330\_F\_0003 DID YOU EVER GO ON STRIKE\\
\ldots{}\\
120958\_20100126\_97016\_M\_0285 SOMETIMES LESS IS BETTER\\
120958\_20100126\_97016\_M\_0286 YOU MUST LOVE TO COOK

Once you've created \texttt{text}, the lexicon will also need to be reduced to only the words present in the corpus. This will ensure that there are no extraneous phones that we are ``training.''

The following code makes a list of words in the corpus and stores it in a file called \texttt{words.txt}. Note that when using the \texttt{cut} command, the default cut is delimited by tab (\texttt{cut\ -f\ 2}), but if the delimiter is anything other than tab, it can be specified as such: \texttt{cut\ -d\ \textquotesingle{}my\ delimiter\textquotesingle{}\ -f\ 2-\ text}. \texttt{words.txt} will serve as input to a script, \texttt{filter\_dict.py}, that downsizes the lexicon to only the words in the corpus.

\begin{Shaded}
\begin{Highlighting}[]
\NormalTok{cut }\OperatorTok{-}\NormalTok{d }\StringTok{' '} \OperatorTok{-}\NormalTok{f }\DecValTok{2}\OperatorTok{-}\StringTok{ }\NormalTok{text }\OperatorTok{|}\StringTok{ }\NormalTok{sed }\StringTok{'s/ /}\CharTok{\textbackslash{}n}\StringTok{/g'} \OperatorTok{|}\StringTok{ }\NormalTok{sort }\OperatorTok{-}\NormalTok{u }\OperatorTok{>}\StringTok{ }\NormalTok{words.txt}
\end{Highlighting}
\end{Shaded}

An example script to accomplish this can be downloaded here: \href{https://www.eleanorchodroff.com/tutorial/kaldi/scripts/filter_dict.py}{filter\_dict.py}

\texttt{filter\_dict.py} takes \texttt{words.txt} and \texttt{lexicon.txt} as input and removes words from the lexicon that are not in the corpus. Remember that \texttt{lexicon.txt} should be in \texttt{/data/local/lang}. You will need to modify the path to \texttt{lexicon.txt} within the script \texttt{filter\_dict.py}. You may also need to change the specified delimiter (tab, comma, space, etc.) within the file. \texttt{filter\_dict.py} returns a modified \texttt{lexicon.txt}.

\begin{Shaded}
\begin{Highlighting}[]
\NormalTok{cd mycorpus}
\NormalTok{python filter_dict.py}
\end{Highlighting}
\end{Shaded}

One more modification needs to be made to the lexicon and that is adding the pseudo-word \textless{}oov\textgreater{} as an entry. \textless{}oov\textgreater{} stands for `out of vocabulary'. Even though we ensured that all words present are indeed in the dictionary, the system requires that this option be present. At the top of your lexicon, add \textless{}oov\textgreater{} \textless{}oov\textgreater{}.

\hypertarget{textfile}{}
\textless{}oov\textgreater{} \textless{}oov\textgreater{}\\
A AH0\\
A EY1

\subsubsection{segments}\label{segments}

The \texttt{segments} file contains the start and end time for each utterance in an audio file. This is a text file with the following format:

\hypertarget{textfile}{}
utt\_id file\_id start\_time end\_time

utt\_id = utterance ID\\
file\_id = file ID\\
start\_time = start time in seconds\\
end\_time = end time in seconds

Example segments file:

\hypertarget{textfile}{}
110236\_20091006\_82330\_F\_001 110236\_20091006\_82330\_F 0.0 3.44\\
110236\_20091006\_82330\_F\_002 110236\_20091006\_82330\_F 4.60 8.54\\
110236\_20091006\_82330\_F\_003 110236\_20091006\_82330\_F 9.45 12.05\\
110236\_20091006\_82330\_F\_004 110236\_20091006\_82330\_F 13.29 16.13\\
110236\_20091006\_82330\_F\_005 110236\_20091006\_82330\_F 17.27 20.36\\
110236\_20091006\_82330\_F\_006 110236\_20091006\_82330\_F 22.06 25.46\\
110236\_20091006\_82330\_F\_007 110236\_20091006\_82330\_F 25.86 27.56\\
110236\_20091006\_82330\_F\_008 110236\_20091006\_82330\_F 28.26 31.24\\
\ldots{}\\
120958\_20100126\_97016\_M\_282 120958\_20100126\_97016\_M 915.62 919.67\\
120958\_20100126\_97016\_M\_283 120958\_20100126\_97016\_M 920.51 922.69\\
120958\_20100126\_97016\_M\_284 120958\_20100126\_97016\_M 922.88 924.27\\
120958\_20100126\_97016\_M\_285 120958\_20100126\_97016\_M 925.35 927.88\\
120958\_20100126\_97016\_M\_286 120958\_20100126\_97016\_M 928.31 930.51

\subsubsection{wav.scp}\label{wav.scp}

\texttt{wav.scp} contains the location for each of the audio files. If your audio files are already in wav format, use the following template:

\hypertarget{textfile}{}
file\_id path/file

Example \texttt{wav.scp} file:

\hypertarget{textfile}{}
110236\_20091006\_82330\_F path/110236\_20091006\_82330\_F.wav\\
111138\_20091215\_82636\_F path/111138\_20091215\_82636\_F.wav\\
111138\_20091217\_82636\_F path/111138\_20091217\_82636\_F.wav\\
\ldots{}\\
120947\_20100125\_59427\_F path/120947\_20100125\_59427\_F.wav\\
120953\_20100125\_79293\_F path/120953\_20100125\_79293\_F.wav\\
120958\_20100126\_97016\_M path/120958\_20100126\_97016\_M.wav

If your audio files are in a different format (sphere, mp3, flac, speex), you will have to convert them to wav format. Instead of having to convert the files manually and storing multiple copies of the data, you can let Kaldi convert the files on-the-fly. The tool \texttt{sox} will come in handy in many of these cases. As an example of sphere (suffix \texttt{.sph}) to wav, you can use the following template; make sure to change \texttt{path} to the actual path where files are located. Also, don't forget the pipe (\texttt{\textbar{}}).

\begin{Shaded}
\begin{Highlighting}[]
\NormalTok{file_id path}\OperatorTok{/}\NormalTok{sph2pipe }\OperatorTok{-}\NormalTok{f wav }\OperatorTok{-}\NormalTok{p }\OperatorTok{-}\NormalTok{c }\DecValTok{1}\NormalTok{ path}\OperatorTok{/}\NormalTok{file }\OperatorTok{|}\StringTok{  }
\end{Highlighting}
\end{Shaded}

For an example using \texttt{sox}, this following code will convert the second channel of an 128kbit/s 44.1kHz joint-stereo mp3 file to a 8kHz mono wav file (which will be processed by Kaldi to generate the features):

\begin{Shaded}
\begin{Highlighting}[]
\NormalTok{file_id path}\OperatorTok{/}\NormalTok{sox audio.mp3 }\OperatorTok{-}\NormalTok{t wav }\OperatorTok{-}\NormalTok{r }\DecValTok{8000} \OperatorTok{-}\NormalTok{c }\DecValTok{1} \OperatorTok{-}\StringTok{ }\NormalTok{remix }\DecValTok{2}\OperatorTok{|}
\end{Highlighting}
\end{Shaded}

\subsubsection{utt2spk}\label{utt2spk}

\texttt{utt2spk} contains the mapping of each utterance to its corresponding speaker. As a side note, engineers will often conflate the term speaker with recording session, such that each recording session is a different ``speaker''. Therefore, the concept of ``speaker'' does not have to be related to a person -- it can be a room, accent, gender, or anything that could influence the recording. When speaker normalization is performed then, the normalization may actually be removing effects due to the recording quality or particular accent type. This definition of ``speaker'' then is left up to the modeler.

\texttt{utt2spk} is a text file with the following format:

\hypertarget{textfile}{}
utt\_id spkr

utt\_id = utterance ID\\
spkr = speaker ID

Example \texttt{utt2spk} file:

\hypertarget{textfile}{}
110236\_20091006\_82330\_F\_0001 110236\\
110236\_20091006\_82330\_F\_0002 110236\\
110236\_20091006\_82330\_F\_0003 110236\\
110236\_20091006\_82330\_F\_0004 110236\\
\ldots{}\\
120958\_20100126\_97016\_M\_0284 120958\\
120958\_20100126\_97016\_M\_0285 120958\\
120958\_20100126\_97016\_M\_0286 120958

Since the speaker ID in the first portion of our utterance IDs, we were able to use the following code to create the \texttt{utt2spk} file:

\begin{Shaded}
\begin{Highlighting}[]
\CommentTok{# this should be interpreted as one line of code}
\NormalTok{cat data}\OperatorTok{/}\NormalTok{train}\OperatorTok{/}\NormalTok{segments }\OperatorTok{|}\StringTok{ }\NormalTok{cut }\OperatorTok{-}\NormalTok{f }\DecValTok{1} \OperatorTok{-}\NormalTok{d }\StringTok{' '} \OperatorTok{|}\StringTok{ }\NormalTok{\textbackslash{}  }
\NormalTok{perl }\OperatorTok{-}\NormalTok{ane }\StringTok{'chomp; @F = split "_", $_; print $_ . " " . @F[0] . "}\CharTok{\textbackslash{}n}\StringTok{";'} \OperatorTok{>}\StringTok{ }\NormalTok{data}\OperatorTok{/}\NormalTok{train}\OperatorTok{/}\NormalTok{utt2spk}
\end{Highlighting}
\end{Shaded}

\subsubsection{spk2utt}\label{spk2utt}

\texttt{spk2utt} is a file that contains the speaker to utterance mapping. This information is already contained in \texttt{utt2spk}, but in the wrong format. The following line of code will automatically create the spk2utt file and simultaneously verify that all data files are present and in the correct format:

\begin{Shaded}
\begin{Highlighting}[]
\NormalTok{utils}\OperatorTok{/}\NormalTok{fix_data_dir.sh data}\OperatorTok{/}\NormalTok{train}
\end{Highlighting}
\end{Shaded}

While \texttt{spk2utt} has already been created, you can verify that it has the following format:

\hypertarget{textfile}{}
spkr utt\_id1 utt\_id2 utt\_id3

\subsection{\texorpdfstring{Create files for \texttt{data/local/lang}}{Create files for data/local/lang}}\label{create-files-for-datalocallang}

\texttt{data/local/lang} is the directory that contains language data specific to the your own corpus. For example, the lexicon only contains words and their pronunciations that are present in the corpus. This directory will contain the following:

\begin{itemize}
\tightlist
\item
  \texttt{lexicon.txt}
\item
  \texttt{nonsilence\_phones.txt}
\item
  \texttt{optional\_silence.txt}
\item
  \texttt{silence\_phones.txt}
\item
  \texttt{extra\_questions.txt} (optional)
\end{itemize}

\subsubsection{lexicon.txt}\label{lexicon.txt}

You will need a pronunciation lexicon of the language you are working on. A good English lexicon is the CMU dictionary, which you can find \href{http://www.speech.cs.cmu.edu/cgi-bin/cmudict}{here}. The lexicon should list each word on its own line, capitalized, followed by its phonemic pronunciation

\hypertarget{textfile}{}
WORD W ER D\\
LEXICON L EH K S IH K AH N

The pronunciation alphabet must be based on the same phonemes you wish to use for your acoustic models. You must also include lexical entries for each ``silence'' or ``out of vocabulary'' phone model you wish to train.

Once you've created the lexicon, move it to \texttt{data/local/lang/}.

\begin{Shaded}
\begin{Highlighting}[]
\NormalTok{cp lexicon.txt kaldi}\OperatorTok{-}\NormalTok{trunk}\OperatorTok{/}\NormalTok{egs}\OperatorTok{/}\NormalTok{mycorpus}\OperatorTok{/}\NormalTok{data}\OperatorTok{/}\NormalTok{local}\OperatorTok{/}\NormalTok{lang}\OperatorTok{/}
\end{Highlighting}
\end{Shaded}

\subsubsection{nonsilence\_phones.txt}\label{nonsilence_phones.txt}

As the name indicates, this file contains a list of all the phones that are not silence. Edit \texttt{phones.txt} so that \emph{like phones} are on the same line. For example, \texttt{AA0}, \texttt{AA1}, and \texttt{AA2} would go on the same line; \texttt{K} would go on a different line. Then save this as \texttt{nonsilence\_phones.txt}.

\begin{Shaded}
\begin{Highlighting}[]
\CommentTok{# this should be interpreted as one line of code}
\NormalTok{cut }\OperatorTok{-}\NormalTok{d }\StringTok{' '} \OperatorTok{-}\NormalTok{f }\DecValTok{2}\OperatorTok{-}\StringTok{ }\NormalTok{lexicon.txt }\OperatorTok{|}\StringTok{  }\NormalTok{\textbackslash{}  }
\NormalTok{sed }\StringTok{'s/ /}\CharTok{\textbackslash{}n}\StringTok{/g'} \OperatorTok{|}\StringTok{ }\NormalTok{\textbackslash{}  }
\NormalTok{sort }\OperatorTok{-}\NormalTok{u }\OperatorTok{>}\StringTok{ }\NormalTok{nonsilence_phones.txt}
\end{Highlighting}
\end{Shaded}

\subsubsection{silence\_phones.txt}\label{silence_phones.txt}

\texttt{silence\_phones.txt} will contain a `SIL' (silence) and `oov' (out of vocabulary) model. optional\_silence.txt will just contain a `SIL' model. This can be created with the following code:

\begin{Shaded}
\begin{Highlighting}[]
\NormalTok{echo –e }\StringTok{'SIL'}\NormalTok{\textbackslash{}\textbackslash{}n}\StringTok{'oov'} \OperatorTok{>}\StringTok{ }\NormalTok{silence_phones.txt}
\end{Highlighting}
\end{Shaded}

\subsubsection{optional\_silence.txt}\label{optional_silence.txt}

\texttt{optional\_silence.txt} will simply contain a `SIL' model. Use the following code to create that file.

\begin{Shaded}
\begin{Highlighting}[]
\NormalTok{echo }\StringTok{'SIL'} \OperatorTok{>}\StringTok{ }\NormalTok{optional_silence.txt}
\end{Highlighting}
\end{Shaded}

\subsubsection{extra\_questions.txt}\label{extra_questions.txt}

A Kaldi script will generate a basic \texttt{extra\_questions.txt} file for you, but in \texttt{data/lang/phones}. This file ``asks questions'' about a phone's contextual information by dividing the phones into two different sets. An algorithm then determines whether it is at all helpful to model that particular context. The standard \texttt{extra\_questions.txt} will contain the most common ``questions.'' An example would be whether the phone is word-initial vs word-final. If you do have extra questions that are not in the standard extra\_questions.txt file, they would need to be added here.

\subsection{\texorpdfstring{Create files for \texttt{data/lang}}{Create files for data/lang}}\label{create-files-for-datalang}

Now that we have all the files in \texttt{data/local/lang}, we can use a script to generate all of the files in \texttt{data/lang}.

\begin{Shaded}
\begin{Highlighting}[]
\NormalTok{cd mycorpus}
\NormalTok{utils}\OperatorTok{/}\NormalTok{prepare_lang.sh data}\OperatorTok{/}\NormalTok{local}\OperatorTok{/}\NormalTok{lang }\StringTok{'OOV'}\NormalTok{ data}\OperatorTok{/}\NormalTok{local}\OperatorTok{/}\StringTok{ }\NormalTok{data}\OperatorTok{/}\NormalTok{lang}
                    
\CommentTok{# where the underlying argument structure is:}
\NormalTok{utils}\OperatorTok{/}\NormalTok{prepare_lang.sh }\OperatorTok{<}\NormalTok{dict}\OperatorTok{-}\NormalTok{src}\OperatorTok{-}\NormalTok{dir}\OperatorTok{>}\StringTok{ }\ErrorTok{<}\NormalTok{oov}\OperatorTok{-}\NormalTok{dict}\OperatorTok{-}\NormalTok{entry}\OperatorTok{>}\StringTok{ }\ErrorTok{<}\NormalTok{tmp}\OperatorTok{-}\NormalTok{dir}\OperatorTok{>}\StringTok{ }\ErrorTok{<}\NormalTok{lang}\OperatorTok{-}\NormalTok{dir}\OperatorTok{>}
\end{Highlighting}
\end{Shaded}

The second argument refers to lexical entry (word) for a ``spoken noise'' or ``out of vocabulary'' phone. Make sure that this entry and its corresponding phone (\texttt{oov}) are entered in \texttt{lexicon.txt} and the phone is listed in \texttt{silence\_phones.txt}.

Note that some older versions of Kaldi allowed the source and tmp directories to refer to the same location. These must now point to different directories.

The new files located in \texttt{data/lang} are \texttt{L.fst}, \texttt{L\_disambig.fst}, \texttt{oov.int}, \texttt{oov.txt}, \texttt{phones.txt}, \texttt{topo}, \texttt{words.txt}, and \texttt{phones}. \texttt{phones} is a directory containing many additional files, including the \texttt{extra\_questions.txt} file mentioned in section \ref{create-files-for-datalocallang}. It is worth taking a look at this file to see how the model may be learning more about a phoneme's contextual information. You should notice fairly logical and linguistically motivated divisions among the phones.

\subsection{Set the parallelization wrapper}\label{set-the-parallelization-wrapper}

Training can be computationally expensive; however, if you have multiple processors/cores or even multiple machines, there are ways to speed it up significantly. Both training and alignment can be made more efficient by splitting the dataset into smaller chunks and processing them in parallel. The number of jobs or splits in the dataset will be specified later in the training and alignment steps. Kaldi provides a wrapper to implement this parallelization so that each of the computational steps can take advantage of the multiple processors. Kaldi's wrapper scripts are \texttt{run.pl}, \texttt{queue.pl}, and \texttt{slurm.pl}, along with a few others we won't discuss here. The applicable script and parameters will then be specified in a file called \texttt{cmd.sh} located at the top level of your corpus' training directory.

\begin{itemize}
\item
  \texttt{run.pl} allows you to run the tasks on a local machine (e.g., your personal computer).
\item
  \texttt{queue.pl} allows you to allocate jobs on machines using \href{https://en.wikipedia.org/wiki/Oracle_Grid_Engine}{Sun Grid Engine} (see also \href{https://en.wikipedia.org/wiki/Grid_computing}{Grid Computing}).
\item
  \texttt{slurm.pl} allows you to allocate jobs on machines using another grid engine software, called \href{https://en.wikipedia.org/wiki/Slurm_Workload_Manager}{SLURM}.
\end{itemize}

The parallelization can be specified separately for training and decoding (alignment of new audio) in the file \texttt{cmd.sh}. The following code provides an example using parameters specific to the Johns Hopkins CLSP cluster. If you are training on a personal computer or do not have a grid engine, you can set \texttt{train\_cmd} and \texttt{decode\_cmd} to \texttt{"run.pl"}.

As a side note, \texttt{vim} is a text editor that operates within the Unix shell. The commented portion of text provides the crucial commands you'll need to know to insert, change modes, write, and quit the editor. Finally, \texttt{cmd.sh} will automatically be created by typing \texttt{vim\ cmd.sh}.

\begin{Shaded}
\begin{Highlighting}[]
\NormalTok{cd mycorpus  }
\NormalTok{vim cmd.sh  }
                    
\CommentTok{# Press i to insert; esc to exit insert mode; }
\CommentTok{# ‘:wq’ to write and quit; ‘:q’ to quit normally; }
\CommentTok{# ‘:q!’ to quit forcibly (without saving)}

\CommentTok{# Insert the following text in cmd.sh}
\NormalTok{train_cmd=}\StringTok{"queue.pl"}
\NormalTok{decode_cmd=}\StringTok{"queue.pl  --mem 2G"}
\end{Highlighting}
\end{Shaded}

Please see \url{http://www.kaldi-asr.org/doc/queue.html} for how to correctly configure this.

Once you've quite vim, then run the file:

\begin{Shaded}
\begin{Highlighting}[]
\NormalTok{cd mycorpus  }
\NormalTok{. .}\OperatorTok{/}\NormalTok{cmd.sh}
\end{Highlighting}
\end{Shaded}

\subsection{\texorpdfstring{Create files for \texttt{conf}}{Create files for conf}}\label{create-files-for-conf}

The directory \texttt{conf} requires one file \texttt{mfcc.conf}, which contains the parameters for MFCC feature extraction. The text file includes the following information:

\hypertarget{textfile}{}
--use-energy=false\\
--sample-frequency=16000

The sampling frequency should be modified to reflect your audio data. This file can be created manually or within the shell with the following code:

\begin{Shaded}
\begin{Highlighting}[]
\CommentTok{# Create mfcc.conf by opening it in a text editor like vim}
\NormalTok{cd mycorpus}\OperatorTok{/}\NormalTok{conf}
\NormalTok{vim mfcc.conf}

\CommentTok{# Press i to insert; esc to exit insert mode; }
\CommentTok{# ‘:wq’ to write and quit; ‘:q’ to quit normally; }
\CommentTok{# ‘:q!’ to quit forcibly (without saving)}

\CommentTok{# Insert the following text in mfcc.conf}
                    
\OperatorTok{--}\NormalTok{use}\OperatorTok{-}\NormalTok{energy=false  }
\OperatorTok{--}\NormalTok{sample}\OperatorTok{-}\NormalTok{frequency=}\DecValTok{16000}
\end{Highlighting}
\end{Shaded}

\subsection{Extract MFCC features}\label{extract-mfcc-features}

The following code will extract the MFCC acoustic features and compute the cepstral mean and variance normalization (CMVN) stats. After each process, it also fixes the data files to ensure that they are still in the correct format.

The \texttt{-\/-nj} option is for the number of jobs to be sent out. This number is currently set to 16 jobs, which means that the data will be divided into 16 sections. It is good to note that Kaldi keeps data from the same speakers together, so you do not want more splits than the number of speakers you have.

\begin{Shaded}
\begin{Highlighting}[]
\NormalTok{cd mycorpus  }
                    
\NormalTok{mfccdir=mfcc  }
\NormalTok{x=data}\OperatorTok{/}\NormalTok{train  }
\NormalTok{steps}\OperatorTok{/}\NormalTok{make_mfcc.sh }\OperatorTok{--}\NormalTok{cmd }\StringTok{"$train_cmd"} \OperatorTok{--}\NormalTok{nj }\DecValTok{16} \OperatorTok{$}\NormalTok{x exp}\OperatorTok{/}\NormalTok{make_mfcc}\OperatorTok{/}\ErrorTok{$}\NormalTok{x }\OperatorTok{$}\NormalTok{mfccdir  }
\NormalTok{steps}\OperatorTok{/}\NormalTok{compute_cmvn_stats.sh }\OperatorTok{$}\NormalTok{x exp}\OperatorTok{/}\NormalTok{make_mfcc}\OperatorTok{/}\ErrorTok{$}\NormalTok{x }\OperatorTok{$}\NormalTok{mfccdir}
\end{Highlighting}
\end{Shaded}

\subsection{Monophone training and alignment}\label{monophone-training-and-alignment}

\begin{itemize}
\tightlist
\item
  \textbf{Take subset of data for monophone training}
\end{itemize}

The monophone models are the first part of the training procedure. We will only train a subset of the data mainly for efficiency. Reasonable monophone models can be obtained with little data, and these models are mainly used to bootstrap training for later models.

The listed argument options for this script indicate that we will take the first part of the dataset, followed by the location the data currently resides in, followed by the number of data points we will take (10,000), followed by the destination directory for the training data.

\begin{Shaded}
\begin{Highlighting}[]
\NormalTok{cd mycorpus  }
\NormalTok{utils}\OperatorTok{/}\NormalTok{subset_data_dir.sh }\OperatorTok{--}\NormalTok{first data}\OperatorTok{/}\NormalTok{train }\DecValTok{10000}\NormalTok{ data}\OperatorTok{/}\NormalTok{train_10k}
\end{Highlighting}
\end{Shaded}

\begin{itemize}
\tightlist
\item
  \textbf{Train monophones}
\end{itemize}

Each of the training scripts takes a similar baseline argument structure with optional arguments preceding those. The one exception is the first monophone training pass. Since a model does not yet exist, there is no source directory specifically for the model. The required arguments are always:

\begin{verbatim}
- Location of the acoustic data: `data/train` 
- Location of the lexicon: `data/lang`
- Source directory for the model: `exp/lastmodel`
- Destination directory for the model: `exp/currentmodel`
\end{verbatim}

The argument \texttt{-\/-cmd\ “\$train\_cmd”} designates which machine should handle the processing. Recall from above that we specified this variable in the file \texttt{cmd.sh}. The argument \texttt{-\/-nj} should be familiar at this point and stands for the number of jobs. Since this is only a subset of the data, we have reduced the number of jobs from 16 to 10. Boost silence is included as standard protocol for this training.

\begin{Shaded}
\begin{Highlighting}[]
\NormalTok{steps}\OperatorTok{/}\NormalTok{train_mono.sh }\OperatorTok{--}\NormalTok{boost}\OperatorTok{-}\NormalTok{silence }\FloatTok{1.25} \OperatorTok{--}\NormalTok{nj }\DecValTok{10} \OperatorTok{--}\NormalTok{cmd }\StringTok{"$train_cmd"}\NormalTok{ \textbackslash{}}
\NormalTok{data}\OperatorTok{/}\NormalTok{train_10k data}\OperatorTok{/}\NormalTok{lang exp}\OperatorTok{/}\NormalTok{mono_10k}
\end{Highlighting}
\end{Shaded}

\begin{itemize}
\tightlist
\item
  \textbf{Align monophones}
\end{itemize}

Just like the training scripts, the alignment scripts also adhere to the same argument structure. The required arguments are always:

\begin{verbatim}
- Location of the acoustic data: `data/train`
- Location of the lexicon: `data/lang`  
- Source directory for the model: `exp/currentmodel`  
- Destination directory for the alignment: `exp/currentmodel_ali`       
\end{verbatim}

\begin{Shaded}
\begin{Highlighting}[]
\NormalTok{steps}\OperatorTok{/}\NormalTok{align_si.sh }\OperatorTok{--}\NormalTok{boost}\OperatorTok{-}\NormalTok{silence }\FloatTok{1.25} \OperatorTok{--}\NormalTok{nj }\DecValTok{16} \OperatorTok{--}\NormalTok{cmd }\StringTok{"$train_cmd"}\NormalTok{ \textbackslash{}}
\NormalTok{data}\OperatorTok{/}\NormalTok{train data}\OperatorTok{/}\NormalTok{lang exp}\OperatorTok{/}\NormalTok{mono_10k exp}\OperatorTok{/}\NormalTok{mono_ali }\OperatorTok{||}\StringTok{ }\NormalTok{exit }\DecValTok{1}\NormalTok{;}
\end{Highlighting}
\end{Shaded}

The directory structure should now look something like this:

\begin{figure}
\centering
\includegraphics{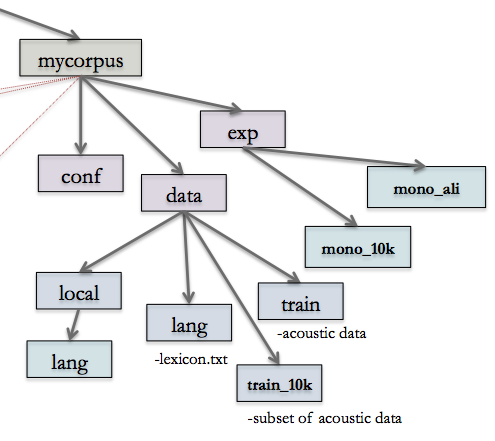}
\caption{Output directory structure}
\end{figure}

\subsection{Triphone training and alignment}\label{triphone-training-and-alignment}

\begin{itemize}
\tightlist
\item
  \textbf{Train delta-based triphones}
\end{itemize}

Training the triphone model includes additional arguments for the number of leaves, or HMM states, on the decision tree and the number of Gaussians. In this command, we specify 2000 HMM states and 10000 Gaussians. As an example of what this means, assume there are 50 phonemes in our lexicon. We could have one HMM state per phoneme, but we know that phonemes will vary considerably depending on if they are at the beginning, middle or end of a word. We would therefore want \emph{at least} three different HMM states for each phoneme. This brings us to a minimum of 150 HMM states to model just that variation. With 2000 HMM states, the model can decide if it may be better to allocate a unique HMM state to more refined allophones of the original phone. This phoneme splitting is decided by the phonetic questions in \texttt{questions.txt} and \texttt{extra\_questions.txt}. The allophones are also referred to as subphones, senones, HMM states, or leaves.

The exact number of leaves and Gaussians is often decided based on heuristics. The numbers will largely depend on the amount of data, number of phonetic questions, and goal of the model. There is also the constraint that the number of Gaussians should always exceed the number of leaves. As you'll see, these numbers increase as we refine our model with further training algorithms.

\begin{Shaded}
\begin{Highlighting}[]
\NormalTok{steps}\OperatorTok{/}\NormalTok{train_deltas.sh }\OperatorTok{--}\NormalTok{boost}\OperatorTok{-}\NormalTok{silence }\FloatTok{1.25} \OperatorTok{--}\NormalTok{cmd }\StringTok{"$train_cmd"}\NormalTok{ \textbackslash{}}
\DecValTok{2000} \DecValTok{10000}\NormalTok{ data}\OperatorTok{/}\NormalTok{train data}\OperatorTok{/}\NormalTok{lang exp}\OperatorTok{/}\NormalTok{mono_ali exp}\OperatorTok{/}\NormalTok{tri1 }\OperatorTok{||}\StringTok{ }\NormalTok{exit }\DecValTok{1}\NormalTok{;}
\end{Highlighting}
\end{Shaded}

\begin{itemize}
\tightlist
\item
  \textbf{Align delta-based triphones}
\end{itemize}

\begin{Shaded}
\begin{Highlighting}[]
\NormalTok{steps}\OperatorTok{/}\NormalTok{align_si.sh }\OperatorTok{--}\NormalTok{nj }\DecValTok{24} \OperatorTok{--}\NormalTok{cmd }\StringTok{"$train_cmd"}\NormalTok{ \textbackslash{}}
\NormalTok{data}\OperatorTok{/}\NormalTok{train data}\OperatorTok{/}\NormalTok{lang exp}\OperatorTok{/}\NormalTok{tri1 exp}\OperatorTok{/}\NormalTok{tri1_ali }\OperatorTok{||}\StringTok{ }\NormalTok{exit }\DecValTok{1}\NormalTok{;}
\end{Highlighting}
\end{Shaded}

\begin{itemize}
\tightlist
\item
  \textbf{Train delta + delta-delta triphones}
\end{itemize}

\begin{Shaded}
\begin{Highlighting}[]
\NormalTok{steps}\OperatorTok{/}\NormalTok{train_deltas.sh }\OperatorTok{--}\NormalTok{cmd }\StringTok{"$train_cmd"}\NormalTok{ \textbackslash{}}
\DecValTok{2500} \DecValTok{15000}\NormalTok{ data}\OperatorTok{/}\NormalTok{train data}\OperatorTok{/}\NormalTok{lang exp}\OperatorTok{/}\NormalTok{tri1_ali exp}\OperatorTok{/}\NormalTok{tri2a }\OperatorTok{||}\StringTok{ }\NormalTok{exit }\DecValTok{1}\NormalTok{;}
\end{Highlighting}
\end{Shaded}

\begin{itemize}
\tightlist
\item
  \textbf{Align delta + delta-delta triphones}
\end{itemize}

\begin{Shaded}
\begin{Highlighting}[]
\NormalTok{steps}\OperatorTok{/}\NormalTok{align_si.sh  }\OperatorTok{--}\NormalTok{nj }\DecValTok{24} \OperatorTok{--}\NormalTok{cmd }\StringTok{"$train_cmd"}\NormalTok{ \textbackslash{}}
\OperatorTok{--}\NormalTok{use}\OperatorTok{-}\NormalTok{graphs true data}\OperatorTok{/}\NormalTok{train data}\OperatorTok{/}\NormalTok{lang exp}\OperatorTok{/}\NormalTok{tri2a exp}\OperatorTok{/}\NormalTok{tri2a_ali  }\OperatorTok{||}\StringTok{ }\NormalTok{exit }\DecValTok{1}\NormalTok{;}
\end{Highlighting}
\end{Shaded}

\begin{itemize}
\tightlist
\item
  \textbf{Train LDA-MLLT triphones}
\end{itemize}

\begin{Shaded}
\begin{Highlighting}[]
\NormalTok{steps}\OperatorTok{/}\NormalTok{train_lda_mllt.sh }\OperatorTok{--}\NormalTok{cmd }\StringTok{"$train_cmd"}\NormalTok{ \textbackslash{}}
\DecValTok{3500} \DecValTok{20000}\NormalTok{ data}\OperatorTok{/}\NormalTok{train data}\OperatorTok{/}\NormalTok{lang exp}\OperatorTok{/}\NormalTok{tri2a_ali exp}\OperatorTok{/}\NormalTok{tri3a }\OperatorTok{||}\StringTok{ }\NormalTok{exit }\DecValTok{1}\NormalTok{;}
\end{Highlighting}
\end{Shaded}

\begin{itemize}
\tightlist
\item
  \textbf{Align LDA-MLLT triphones with FMLLR}
\end{itemize}

\begin{Shaded}
\begin{Highlighting}[]
\NormalTok{steps}\OperatorTok{/}\NormalTok{align_fmllr.sh }\OperatorTok{--}\NormalTok{nj }\DecValTok{32} \OperatorTok{--}\NormalTok{cmd }\StringTok{"$train_cmd"}\NormalTok{ \textbackslash{}}
\NormalTok{data}\OperatorTok{/}\NormalTok{train data}\OperatorTok{/}\NormalTok{lang exp}\OperatorTok{/}\NormalTok{tri3a exp}\OperatorTok{/}\NormalTok{tri3a_ali }\OperatorTok{||}\StringTok{ }\NormalTok{exit }\DecValTok{1}\NormalTok{;}
\end{Highlighting}
\end{Shaded}

\begin{itemize}
\tightlist
\item
  \textbf{Train SAT triphones}
\end{itemize}

\begin{Shaded}
\begin{Highlighting}[]
\NormalTok{steps}\OperatorTok{/}\NormalTok{train_sat.sh  }\OperatorTok{--}\NormalTok{cmd }\StringTok{"$train_cmd"}\NormalTok{ \textbackslash{}}
\DecValTok{4200} \DecValTok{40000}\NormalTok{ data}\OperatorTok{/}\NormalTok{train data}\OperatorTok{/}\NormalTok{lang exp}\OperatorTok{/}\NormalTok{tri3a_ali exp}\OperatorTok{/}\NormalTok{tri4a }\OperatorTok{||}\StringTok{ }\NormalTok{exit }\DecValTok{1}\NormalTok{;}
\end{Highlighting}
\end{Shaded}

\begin{itemize}
\tightlist
\item
  \textbf{Align SAT triphones with FMLLR}
\end{itemize}

\begin{Shaded}
\begin{Highlighting}[]
\NormalTok{steps}\OperatorTok{/}\NormalTok{align_fmllr.sh  }\OperatorTok{--}\NormalTok{cmd }\StringTok{"$train_cmd"}\NormalTok{ \textbackslash{}}
\NormalTok{data}\OperatorTok{/}\NormalTok{train data}\OperatorTok{/}\NormalTok{lang exp}\OperatorTok{/}\NormalTok{tri4a exp}\OperatorTok{/}\NormalTok{tri4a_ali }\OperatorTok{||}\StringTok{ }\NormalTok{exit }\DecValTok{1}\NormalTok{;}
\end{Highlighting}
\end{Shaded}

\section{Forced Alignment}\label{forced-alignment}

Once acoustic models have been created, Kaldi can also perform forced alignment on audio accompanied by a word-level transcript. Note that the \href{https://montreal-forced-aligner.readthedocs.io/en/latest/}{Montreal Forced Aligner} is a forced alignment system based on Kaldi-trained acoustic models for several world languages. You could also considering checking out \href{https://github.com/JoFrhwld/FAVE/wiki/FAVE-align}{FAVE} for aligning American English speech.

Otherwise, if the audio to be aligned is the same as the audio used in the acoustic models, then the alignments can be extracted directly from the alignment files. If you have new audio and transcripts, then the transcript files will need to be updated before alignment.

The full procedure will convert output from the model alignment into Praat TextGrids containing the phone-level transcript.

If the data to be aligned is the same as the training data, skip to Section \ref{extract-alignment}. Otherwise, you'll need to update the transcript files and audio file specifications.

\subsection{Prepare alignment files}\label{prepare-alignment-files}

To extract alignments for new transcripts and audio, you'll need to create new versions of the files in the directory \texttt{data/train}. As a reminder, these files are \texttt{text}, \texttt{segments}, \texttt{wav.scp}, \texttt{utt2spk}, and \texttt{spk2utt} (see Section \ref{create-files-for-datatrain}). We'll house these in a new directory in \texttt{mycorpus/data}.

\begin{Shaded}
\begin{Highlighting}[]
\CommentTok{# create text, segments, wav.scp, utt2spk, and spk2utt}

\NormalTok{cd mycorpus}\OperatorTok{/}\NormalTok{data}
\NormalTok{mkdir alignme         }
\end{Highlighting}
\end{Shaded}

\subsection{Extract MFCC features}\label{extract-mfcc-features-1}

Revisit Section \ref{extract-mfcc-features} on MFCC feature extraction for reference. You'll need to replace \texttt{data/train} with the the new directory, \texttt{data/alignme}.

\begin{Shaded}
\begin{Highlighting}[]
\NormalTok{cd mycorpus}
                    
\NormalTok{mfccdir=mfcc}
\ControlFlowTok{for}\NormalTok{ x }\ControlFlowTok{in}\NormalTok{ data}\OperatorTok{/}\NormalTok{alignme}
\NormalTok{do}
\NormalTok{    steps}\OperatorTok{/}\NormalTok{make_mfcc.sh }\OperatorTok{--}\NormalTok{cmd }\StringTok{"$train_cmd"} \OperatorTok{--}\NormalTok{nj }\DecValTok{16} \OperatorTok{$}\NormalTok{x exp}\OperatorTok{/}\NormalTok{make_mfcc}\OperatorTok{/}\ErrorTok{$}\NormalTok{x }\OperatorTok{$}\NormalTok{mfccdir}
\NormalTok{    utils}\OperatorTok{/}\NormalTok{fix_data_dir.sh data}\OperatorTok{/}\NormalTok{alignme}
\NormalTok{    steps}\OperatorTok{/}\NormalTok{compute_cmvn_stats.sh }\OperatorTok{$}\NormalTok{x exp}\OperatorTok{/}\NormalTok{make_mfcc}\OperatorTok{/}\ErrorTok{$}\NormalTok{x }\OperatorTok{$}\NormalTok{mfccdir}
\NormalTok{    utils}\OperatorTok{/}\NormalTok{fix_data_dir.sh data}\OperatorTok{/}\NormalTok{alignme}
\NormalTok{done}
\end{Highlighting}
\end{Shaded}

\subsection{Align data}\label{align-data}

Revisit Section \ref{triphone-training-and-alignment} on triphone training and alignment for reference. Select the acoustic model and corresponding alignment process you'd like to use. You'll need to replace \texttt{data/train} with the the new directory, \texttt{data/alignme}. As an example:

\begin{Shaded}
\begin{Highlighting}[]
\NormalTok{cd mycorpus}
\NormalTok{steps}\OperatorTok{/}\NormalTok{align_si.sh }\OperatorTok{--}\NormalTok{cmd }\StringTok{"$train_cmd"}\NormalTok{ data}\OperatorTok{/}\NormalTok{alignme data}\OperatorTok{/}\NormalTok{lang \textbackslash{}}
\NormalTok{exp}\OperatorTok{/}\NormalTok{tri4a exp}\OperatorTok{/}\NormalTok{tri4a_alignme }\OperatorTok{||}\StringTok{ }\NormalTok{exit }\DecValTok{1}\NormalTok{;}
\end{Highlighting}
\end{Shaded}

\subsection{Extract alignment}\label{extract-alignment}

\begin{itemize}
\tightlist
\item
  \textbf{Obtain CTM output from alignment files}
\end{itemize}

CTM stands for time-marked conversation file and contains a time-aligned phoneme transcription of the utterances. Its format is:

\hypertarget{textfile}{}
utt\_id channel\_num start\_time phone\_dur phone\_id

To obtain these, you will need to decide which acoustic models to use. The following code will extract the CTM output from the alignment files in the directory \texttt{tri4a\_alignme}, using the acoustic models in \texttt{tri4a}:

\begin{Shaded}
\begin{Highlighting}[]
\NormalTok{cd mycorpus}
                    
\ControlFlowTok{for}\NormalTok{ i }\ControlFlowTok{in}\NormalTok{  exp}\OperatorTok{/}\NormalTok{tri4a_alignme}\OperatorTok{/}\NormalTok{ali.}\OperatorTok{*}\NormalTok{.gz;}
\NormalTok{do src}\OperatorTok{/}\NormalTok{bin}\OperatorTok{/}\NormalTok{ali}\OperatorTok{-}\NormalTok{to}\OperatorTok{-}\NormalTok{phones }\OperatorTok{--}\NormalTok{ctm}\OperatorTok{-}\NormalTok{output exp}\OperatorTok{/}\NormalTok{tri4a}\OperatorTok{/}\NormalTok{final.mdl \textbackslash{}}
\NormalTok{ark}\OperatorTok{:}\StringTok{"gunzip -c $i|"}\NormalTok{ ->}\StringTok{ }\ErrorTok{$}\NormalTok{\{i%.gz\}.ctm;}
\NormalTok{done;}
\end{Highlighting}
\end{Shaded}

\begin{itemize}
\tightlist
\item
  \textbf{Concatenate CTM files}
\end{itemize}

\begin{Shaded}
\begin{Highlighting}[]
\NormalTok{cd mycorpus}\OperatorTok{/}\NormalTok{exp}\OperatorTok{/}\NormalTok{tri4a_alignme}
\NormalTok{cat }\OperatorTok{*}\NormalTok{.ctm }\OperatorTok{>}\StringTok{ }\NormalTok{merged_alignment.txt}
\end{Highlighting}
\end{Shaded}

\begin{itemize}
\tightlist
\item
  \textbf{Convert time marks and phone IDs}
\end{itemize}

The CTM output reports start and end times relative to the utterance, as opposed to the file. You will need the \texttt{segments} file located in either \texttt{data/train} or \texttt{data/alignme} to convert the utterance times into file times.

The output also reports the phone ID, as opposed to the phone itself. You will need the \texttt{phones.txt} file located in \texttt{data/lang} to convert the phone IDs into phone symbols.

An example script to accommplish this can be downloaded here: \href{https://www.eleanorchodroff.com/tutorial/kaldi/scripts/id2phone.R}{id2phone.R}

After obtaining the \texttt{segments} and \texttt{phones.txt} files, run \texttt{id2phone.R} to convert phone IDs to phones characters and map utterance times to file times. You will need to modify the file locations and possibly the regular expression to obtain the filename from the utterance name. Recall that the CTM output lists the utterance ID whereas the segments file lists the file ID. (If you named things logically, the file ID should be a subset of the utterance ID.)

\texttt{id2phone.R} returns a modified version of \texttt{merged\_alignment.txt} called \texttt{final\_ali.txt}

\begin{itemize}
\tightlist
\item
  \textbf{Split \texttt{final\_ali.txt} by file}
\end{itemize}

An example script to accomplish this can be downloaded here: \href{https://www.eleanorchodroff.com/tutorial/kaldi/scripts/splitAlignments.py}{splitAlignments.py}

\texttt{final\_ali.txt} contains the phone transcript for all files together. This can be split into unique files by running \texttt{splitAlignments.py}. You will need to modify the location of \texttt{final\_ali.txt} in this script.

\begin{Shaded}
\begin{Highlighting}[]
\NormalTok{python splitAlignments.py}
\end{Highlighting}
\end{Shaded}

\begin{itemize}
\tightlist
\item
  \textbf{Create word alignments from phone endings}
\end{itemize}

First we'll need to use the {[}B I E S{]} suffixes on the phones in order to group phones together into word-level units.

Run \href{https://www.eleanorchodroff.com/tutorial/kaldi/scripts/phons2pron.py}{phons2pron.py} to complete this step. Note that I have utf-8 character encoding on this script. If necessary, this can be updated to reflect the character encoding that best matches your files.

Second, we'll need to match the phone pronunciation to the corresponding lexical entry using \texttt{lexicon.txt}.

Run \href{https://www.eleanorchodroff.com/tutorial/kaldi/scripts/pron2words.py}{pron2words.py} to complete this step.

\subsection{Create Praat TextGrids}\label{create-praat-textgrids}

\begin{itemize}
\tightlist
\item
  \textbf{Append header to each of the text files for Praat}
\end{itemize}

Praat requires that a text file have a header. Once we append the header, then we can convert these text files into TextGrids. The following code requires a text file containing the header:

\hypertarget{textfile}{}
file\_utt file id ali startinutt dur phone start\_utt end\_utt start end

It also requires a \texttt{tmp} directory for processing. I put this on my Desktop.

\begin{Shaded}
\begin{Highlighting}[]
\NormalTok{cd }\OperatorTok{~}\ErrorTok{/}\NormalTok{Desktop}
\NormalTok{mkdir tmp}
                
\NormalTok{header=}\StringTok{"/Users/Eleanor/Desktop/header.txt"}
                
\CommentTok{# direct the terminal to the directory with the newly split session files}
\CommentTok{# ensure that the RegEx below will capture only the session files}
\CommentTok{# otherwise change this or move the other .txt files to a different folder}
                
\NormalTok{cd mycorpus}\OperatorTok{/}\NormalTok{forcedalignment}
\ControlFlowTok{for}\NormalTok{ i }\ControlFlowTok{in} \OperatorTok{*}\NormalTok{.txt;}
\NormalTok{do}
\NormalTok{    cat }\StringTok{"$header"} \StringTok{"$i"} \OperatorTok{>}\StringTok{ }\ErrorTok{/}\NormalTok{Users}\OperatorTok{/}\NormalTok{Eleanor}\OperatorTok{/}\NormalTok{Desktop}\OperatorTok{/}\NormalTok{tmp}\OperatorTok{/}\NormalTok{xx.}\OperatorTok{$}\ErrorTok{$}
\StringTok{    }\NormalTok{mv }\OperatorTok{/}\NormalTok{Users}\OperatorTok{/}\NormalTok{Eleanor}\OperatorTok{/}\NormalTok{Desktop}\OperatorTok{/}\NormalTok{tmp}\OperatorTok{/}\NormalTok{xx.}\OperatorTok{$}\ErrorTok{$}\StringTok{ "$i"}
\NormalTok{done;}
\end{Highlighting}
\end{Shaded}

\begin{itemize}
\tightlist
\item
  \textbf{Make Praat TextGrids of phone alignments from \texttt{.txt} files}
\end{itemize}

\href{https://www.eleanorchodroff.com/tutorial/kaldi/scripts/createtextgrid.praat}{createtextgrid.praat} will read in the new phone transcripts and corresponding audio files to create a TextGrid for that file. You will need to modify the locations of the phone transcripts and audio files.

\begin{itemize}
\tightlist
\item
  \textbf{Make Praat TextGrids for word alignments from \texttt{word\_alignment.txt}}
\end{itemize}

An example script to accomplish this can be downloaded here: \href{https://www.eleanorchodroff.com/tutorial/kaldi/scripts/createWordTextGrids.praat}{createWordTextGrids.praat}

\begin{itemize}
\tightlist
\item
  \textbf{Stack phone and word TextGrids}
\end{itemize}

\href{https://www.eleanorchodroff.com/tutorial/kaldi/scripts/stackTextGrids.praat}{stackTextGrids.praat}

\chapter{FAVE-align}\label{fave-align}

\section{Overview}\label{overview-1}

\textbf{What does FAVE do?} FAVE is up-to-date implementation of the Penn Forced Aligner covered in section \ref{penn-forced-aligner-legacy}. Two programs are included in installation: FAVE-align, which performs forced alignment on American English speech, and FAVE-extract, which performs formant extraction and normalization algorithms. This tutorial focuses on FAVE-align, so future references of FAVE refer to the alignment program only. FAVE takes a sound file and utterance-level transcript (text file) and returns a Praat TextGrid with phone and word alignments. Excellent documentation and instructions for running the aligner can be found on the \href{https://github.com/JoFrhwld/FAVE/wiki/FAVE-align}{FAVE-align website}. I will cover many of the instructions again, and try to include some additional tips, but provided installation goes smoothly, this aligner tends to work very well right out of the box.

\textbf{What does it include?} As in the Penn Forced Aligner, the system comes with \emph{pre-trained acoustic models} of American English (AE) speech from 25 hours of the SCOTUS corpus, built with the HTK Speech Recognition Toolkit. The section on \href{https://www.eleanorchodroff.com/tutorial/kaldi/introduction.html}{Kaldi} introduces a similar, but alternative system to HTK, in case you'd like to train your own acoustic models from scratch. In addition to the acoustic models, the download also includes a large lexicon of words based on the CMU Pronouncing Dictionary. The dictionary contains over 100,000 words with their standard pronunciation transcriptions in \href{http://en.wikipedia.org/wiki/Arpabet}{Arpabet}. Arpabet is a machine-readable phonetic alphabet of General American English with stress marked on the vowels. While I have, perhaps wrongly, used the Penn Forced Aligner to create phonetic alignments for other languages, those transcriptions had to be forced into Arpabet phones. For a quick alignment, this works fine, but please be advised: the acoustic models of the Penn Forced Aligner are trained on American English, so any speech it is given will be treated like American English. Unless manual adjustments follow the automatic alignment, the PFA alignment would not be ideal for most non-AE phonetic analyses. See the Montreal Forced Aligner (section \ref{montreal-forced-aligner}) for pre-trained acoustic models of additional languages and \href{https://www.eleanorchodroff.com/tutorial/kaldi/introduction.html}{Kaldi} if you would like to create custom acoustic models for a particular language or dialect.

\section{Installation}\label{installation-1}

Please refer to the \href{https://github.com/JoFrhwld/FAVE/wiki/Installing-FAVE-align}{FAVE website} for installation. As of writing, that page currently includes links to OS-specific downloads for each of the prerequisites.

The prerequisites for FAVE are:

\begin{itemize}
\tightlist
\item
  HTK Toolkit Version 3.4.1

  \begin{itemize}
  \tightlist
  \item
    Unlike p2fa, version 3.4.1 \textbf{will} work
  \item
    If installing HTK on a Mac, there are additional prerequisites you'll need. Make sure to check out the \href{https://github.com/JoFrhwld/FAVE/wiki/HTK-on-OS-X}{FAVE wiki} for these instructions.
  \end{itemize}
\item
  Python 2.x

  \begin{itemize}
  \tightlist
  \item
    Pre-installed on Macs, but you can double-check by opening the terminal and typing python. This will return the version installed. I have been using Version 2.7 and have not had any issues running the aligner. Since you've typed python, the terminal is now running with the assumption of python syntax. You want to exit this to get back to its Unix base; to do this, type \texttt{exit()} or \texttt{quit()}. If you have an earlier version of Python, you may need to type \texttt{exit}.
  \end{itemize}
\item
  SoX

  \begin{itemize}
  \tightlist
  \item
    Download here: \url{http://sourceforge.net/projects/sox/}
  \end{itemize}
\end{itemize}

To download FAVE, again follow the instructions on the website. FAVE can either be downloaded directly as a .zip or .tar.gz file \href{https://github.com/JoFrhwld/FAVE/releases}{here} or cloned via git using the following command:

\begin{Shaded}
\begin{Highlighting}[]
\NormalTok{git clone https}\OperatorTok{:}\ErrorTok{//}\NormalTok{github.com}\OperatorTok{/}\NormalTok{JoFrhwld}\OperatorTok{/}\NormalTok{FAVE.git}
\end{Highlighting}
\end{Shaded}

Cloning the system ensures that any new updates are automatically applied to the system.

\section{Running the aligner}\label{running-the-aligner}

FAVE requires a wav file and a corresponding transcript that must adhere to a precise format, which is described below. Unlike some other alignment systems, the wav file does not need to have a sampling rate of 16 kHz. During alignment, the system will automatically create a temporary file with this sampling rate.

The text file must be tab-delimited (.txt) with the following five columns:

\begin{enumerate}
\def\labelenumi{\arabic{enumi}.}
\tightlist
\item
  Speaker ID
\item
  Speaker name
\item
  Onset time (seconds)
\item
  Offset time (seconds)
\item
  Transcription of speech between onset and offset times
\end{enumerate}

Some notes I've taken that could be useful:

\begin{itemize}
\item
  The speaker ID can be the same as the speaker name
\item
  You can make up a very large number as a hack for indicating that the offset time should be the end of the file. The aligner will produce a warning, but still complete the alignment.
\item
  FAVE sometimes struggles with spontaneous speech, though I imagine any forced alignment system might struggle with this. When the total number of phones in the transcript exceeded the total utterance time under the assumption that a phone is 30 ms, then the aligner produced overlapping intervals. This resulted in areas of Praat TextGrids with overlapping intervals which are visible but no longer functional, and now somewhat useless. It could be a bug in either Praat or FAVE.
\item
  Some people have asked me how to align multiple files at once. This can be accomplished with a for loop in the shell script that uses regular expression matching to loop through files. Assuming the transcripts and wav files are in the same folder and differ only in their extension (.txt vs .wav), then you could use the following code (make sure to remove the backslashes and run as one line of code):
\end{itemize}

\begin{Shaded}
\begin{Highlighting}[]
\CommentTok{# direct shell to location of FAAValign.py script}
\NormalTok{cd }\OperatorTok{/}\NormalTok{Users}\OperatorTok{/}\NormalTok{Eleanor}\OperatorTok{/}\NormalTok{FAVE}\OperatorTok{/}\NormalTok{FAVE}\OperatorTok{-}\NormalTok{align}

\CommentTok{# loop through transcript files in a second directory that contains }
\CommentTok{# the transcript files and like-named wav files}
\ControlFlowTok{for}\NormalTok{ i }\ControlFlowTok{in} \OperatorTok{/}\NormalTok{Users}\OperatorTok{/}\NormalTok{Eleanor}\OperatorTok{/}\NormalTok{myCorpus}\OperatorTok{/}\ErrorTok{*}\NormalTok{.txt; \textbackslash{}}
\NormalTok{do python FAAValign.py }\StringTok{"$\{i/.txt/.wav\}"} \StringTok{"$i"} \StringTok{"$\{i/.txt/.TextGrid\}"}\NormalTok{; \textbackslash{}}
\NormalTok{done;}
\end{Highlighting}
\end{Shaded}

\chapter{Montreal Forced Aligner}\label{montreal-forced-aligner}

\section{Overview}\label{overview-2}

The \href{https://montreal-forced-aligner.readthedocs.io/en/latest/introduction.html}{Montreal Forced Aligner} is a forced alignment system with acoustic models built using the \href{https://eleanorchodroff.com/tutorial/kaldi/introduction.html}{Kaldi ASR toolkit}. A major highlight of this system is the availability of pretrained acoustic models and grapheme-to-phoneme models for a wide variety of languages.

The primary website contains excellent documentation, so I'll provide some tips and tricks I've picked up while using it.

A quick link to the installation instructions is located on the primary MFA \href{https://montreal-forced-aligner.readthedocs.io/en/latest/installation.html}{website}. This tutorial is based on Version 1.1.

\section{Setup}\label{setup}

As with any forced alignment system, the Montreal Forced Aligner will time-align a transcript to a corresponding audio file at the phone and word levels provided there exist a set of pretrained acoustic models and a lexicon/dictionary of the words in the transcript with their canonical phonetic pronunciation(s). The phone set used in the dictionary must match the phone set in the acoustic models. The orthography used in the dictionary must also match that in the transcript.

Very generally, the procedure is as follows:

\begin{itemize}
\tightlist
\item
  Prep wav file(s) (16 kHz, single channel)
\item
  Prep transcript(s) (Praat TextGrid or .lab/.txt file)
\item
  Obtain a pronunciation lexicon
\item
  Obtain acoustic models
\end{itemize}

You will also need to identify or create an \textbf{input folder} that contains the wav files and TextGrids/transcripts and an \textbf{output folder} for the time-aligned TextGrid to be created.

\textbf{Please make sure that you have separate input and output folders, and that the output folder is not a subdirectory of the input folder!} The MFA deletes everything in the output folder: if it is the same as your input folder, the system will delete your input files.

\subsection{Wav files}\label{wav-files}

The Montreal Forced Aligner works best and sometimes will only work with wav files that are sampled at 16 kHz and are single channel files. You may need to resample your audio and extract a single channel prior to running the aligner.

\href{https://www.eleanorchodroff.com/tutorial/scripts/prep_audio_mfa.praat}{prep\_audio\_mfa.praat}

\subsection{Transcripts}\label{transcripts}

The MFA can take as input either a Praat TextGrid or a \texttt{.lab} or \texttt{.txt} file. I have worked most extensively with the TextGrid input, so I'll describe those details here. As for \texttt{.lab} and \texttt{.txt} input, I have only tried running the aligner where the transcript is pasted in as a single line. I think there is a way of providing timestamps at the utterance level, but I can't speak to that yet.

The most straightforward implementation of the aligner with TextGrid input is to paste the transcript into a TextGrid with a single interval tier. The transcript \emph{must} be delimited by boundaries on that tier; however, those boundaries \emph{cannot} be located at either the absolute start or absolute end of the wav file (start boundary != 0, end boundary != total duration). In fact, I've found that the MFA can be very sensitive to the location of the end boundary: it's best to have at least 20 ms, if not 50 ms+ between the final TextGrid boundary and the end of the wav file (see also Section \ref{tips-and-tricks} on Tips and Tricks).

If you have utterance-level timestamps, you can also add in additional intervals for an alignment that is less likely to ``derail''. By ``derail'', I mean that the aligner gets thrown off early on in the wav file and never gets back on track, which yields a fairly misaligned ``alignment''. By delimiting the temporal span of an utterance, the aligner has a chance to reset at the next utterance, even if the preceding utterance was completely misaligned. Side note: misalignments are more likely to occur if there's additional noise in the wav file (e.g., coughing, background noise) or if the speech and transcript don't match at either the word or phone level (e.g., pronunciation of a word does not match the dictionary/lexicon entry).

Here are few sample Praat scripts I employ for creating TextGrids.

If I don't have timestamps, but I do have a transcript: \href{https://www.eleanorchodroff.com/tutorial/scripts/create_textgrid_mfa_simple.praat}{create\_textgrid\_mfa\_simple.praat}

If the transcript has start and end times for each utterance (3 column text file with start time, end time, text): \href{https://www.eleanorchodroff.com/tutorial/scripts/create_textgrid_mfa_timestamps.praat}{create\_textgrid\_mfa\_timestamps.praat}

That last Praat script can also be modified for a transcript with either start or end times, but not both. Make sure to follow the ``rules'' (which may change) that text-containing intervals be separated by empty intervals and the boundaries do not align with either the absolute start or end of the file.

\subsection{Pronunciation lexicon}\label{pronunciation-lexicon}

The pronunciation lexicon must be a two column text file with a list of words on the lefthand side and the phonetic pronunciation(s) on the righthand side. Many-to-many mappings between words and pronunciations are permitted. As mentioned above, the phone set must match that used in the acoustic models and the orthography must match that in the transcripts.

There are a few options for obtaining a pronunciation lexicon, outlined below. More details about several of these options are in the sections to come.

\begin{itemize}
\item
  \href{https://montreal-forced-aligner.readthedocs.io/en/latest/pretrained_models.html}{Download} the pronunciation lexicon from the MFA website

  \begin{itemize}
  \tightlist
  \item
    As of writing, there are dictionaries for English, French, and German
  \end{itemize}
\item
  \href{https://montreal-forced-aligner.readthedocs.io/en/latest/pretrained_models.html}{Generate} the pronunciation lexicon from the transcripts using a pretrained grapheme-to-phoneme (G2P) model

  \begin{itemize}
  \tightlist
  \item
    See section \ref{grapheme-to-phoneme-models} on Running a G2P model
  \end{itemize}
\item
  \href{https://montreal-forced-aligner.readthedocs.io/en/latest/g2p_model_training.html}{Train} a G2P model to then generate the pronunciation lexicon
\item
  Create the pronunciation lexicon by hand using the same phone set as the acoustic models
\end{itemize}

\subsection{Acoustic models}\label{acoustic-models}

Pretrained acoustic models for several languages can be downloaded from the Montreal Forced Aligner website.

If you wish to train custom acoustic models on a speech corpus, this can be accomplished using the Kaldi ASR toolkit. A tutorial for training acoustic models can be found \href{https://eleanorchodroff.com/tutorial/kaldi/introduction.html}{here}.

\section{Grapheme-to-phoneme models}\label{grapheme-to-phoneme-models}

If you need a lexicon for the words in your transcript, you might be able to generate one using a grapheme-to-phoneme model. Grapheme-to-phoneme models convert the orthographic representation of a language to its canonical phonetic form after having been trained on examples or conversion rules. Pretrained grapheme-to-phoneme (G2P) models can be found at the \href{https://montreal-forced-aligner.readthedocs.io/en/latest/pretrained_models.html\#pretrained-g2p-models}{Montreal Forced Aligner website}. Once you download the one you want, you can follow these instructions:

\begin{enumerate}
\def\labelenumi{\arabic{enumi}.}
\item
  Place grapheme-to-phoneme model in \texttt{montreal-forced-aligner/pretrained\_models} folder (they can technically go anywhere, but this structure keeps the files organized)
\item
  Create input and output folders
\item
  Place transcripts and wav files in input folder. At least in version 1.1, the wav files needed to be present in order to run the grapheme-to-phoneme conversion model on the transcripts
\item
  Run grapheme-to-phoneme model
\end{enumerate}

\begin{Shaded}
\begin{Highlighting}[]
\NormalTok{cd path}\OperatorTok{/}\NormalTok{to}\OperatorTok{/}\NormalTok{montreal}\OperatorTok{-}\NormalTok{forced}\OperatorTok{-}\NormalTok{aligner}\OperatorTok{/}

\NormalTok{bin}\OperatorTok{/}\NormalTok{mfa_generate_dictionary }\OperatorTok{/}\NormalTok{path}\OperatorTok{/}\NormalTok{to}\OperatorTok{/}\NormalTok{model}\OperatorTok{/}\NormalTok{file.zip }\OperatorTok{/}\NormalTok{path}\OperatorTok{/}\NormalTok{to}\OperatorTok{/}\NormalTok{corpus }\OperatorTok{/}\NormalTok{path}\OperatorTok{/}\NormalTok{to}\OperatorTok{/}\NormalTok{save.txt}
\end{Highlighting}
\end{Shaded}

\texttt{bin/mfa\_generate\_dictionary} takes 3 arguments:

\begin{verbatim}
1. where is the grapheme-to-phoneme model?  
2. where are the wav files and transcripts? (input folder)  
3. where should the output go? (output text file)
\end{verbatim}

Explicit example (make sure to remove backslashes):

\begin{Shaded}
\begin{Highlighting}[]
\NormalTok{cd }\OperatorTok{/}\NormalTok{Users}\OperatorTok{/}\NormalTok{Eleanor}\OperatorTok{/}\NormalTok{montreal}\OperatorTok{-}\NormalTok{forced}\OperatorTok{-}\NormalTok{aligner}

\NormalTok{bin}\OperatorTok{/}\NormalTok{mfa_generate_dictionary pretrained_models}\OperatorTok{/}\NormalTok{mandarin_character_g2p.zip \textbackslash{}}
\OperatorTok{/}\NormalTok{Users}\OperatorTok{/}\NormalTok{Eleanor}\OperatorTok{/}\NormalTok{Desktop}\OperatorTok{/}\NormalTok{align_input }\OperatorTok{/}\NormalTok{Users}\OperatorTok{/}\NormalTok{Eleanor}\OperatorTok{/}\NormalTok{Desktop}\OperatorTok{/}\NormalTok{mandarin_dict.txt}
\end{Highlighting}
\end{Shaded}

\section{Running the aligner}\label{running-the-aligner-1}

\begin{enumerate}
\def\labelenumi{\arabic{enumi}.}
\item
  Place acoustic models and dictionary in \texttt{montreal-forced-aligner/pretrained\_models} folder (they can technically go anywhere, but this structure keeps the files organized)
\item
  Create input and output folders
\item
  Place TextGrids and wav files in input folder
\item
  Run Montreal Forced Aligner
\end{enumerate}

Make sure to change the arguments of \texttt{bin/mfa\_align}!

\begin{Shaded}
\begin{Highlighting}[]
\NormalTok{cd path}\OperatorTok{/}\NormalTok{to}\OperatorTok{/}\NormalTok{montreal}\OperatorTok{-}\NormalTok{forced}\OperatorTok{-}\NormalTok{aligner}\OperatorTok{/}

\NormalTok{bin}\OperatorTok{/}\NormalTok{mfa_align corpus_directory dictionary acoustic_model output_directory}
\end{Highlighting}
\end{Shaded}

\texttt{bin/mfa\_align} takes 4 arguments:

\begin{verbatim}
1. where are the wav files and TextGrids? (input folder)  
2. where is the dictionary?  
3. where are the acoustic models? (you do need the .zip extension)  
4. where should the output go? (output folder)  
\end{verbatim}

Explicit example (make sure to remove backslashes):

\begin{Shaded}
\begin{Highlighting}[]
\NormalTok{cd }\OperatorTok{/}\NormalTok{Users}\OperatorTok{/}\NormalTok{Eleanor}\OperatorTok{/}\NormalTok{montreal}\OperatorTok{-}\NormalTok{forced}\OperatorTok{-}\NormalTok{aligner}

\NormalTok{bin}\OperatorTok{/}\NormalTok{mfa_align }\OperatorTok{/}\NormalTok{Users}\OperatorTok{/}\NormalTok{Eleanor}\OperatorTok{/}\NormalTok{Desktop}\OperatorTok{/}\NormalTok{align_input pretrained_models}\OperatorTok{/}\NormalTok{german_dictionary.txt \textbackslash{}}
\NormalTok{pretrained_models}\OperatorTok{/}\NormalTok{german.zip }\OperatorTok{/}\NormalTok{Users}\OperatorTok{/}\NormalTok{Eleanor}\OperatorTok{/}\NormalTok{Desktop}\OperatorTok{/}\NormalTok{align_output}
\end{Highlighting}
\end{Shaded}

\section{Tips and tricks}\label{tips-and-tricks}

\textbf{Acoustic models}

You do not need to unzip these. If you do, make sure to call the \texttt{.zip} version.

\textbf{Wav files}

I mentioned it above, and will mention it again. Things tend to go more smoothly when the wav file is already 16 kHz and a single channel.

\textbf{TextGrids}

\begin{itemize}
\tightlist
\item
  make sure TextGrid boundaries do not align with either the absolute start or end of the file\\
\item
  make sure the final TextGrid boundary is at least \textasciitilde{}20-50 ms away from the edge (if it still doesn't work, you might want to try increasing that interval)
\item
  sometimes it helps to have an empty interval between every interval containing text\\
\item
  sometimes it helps to increase the number of intervals present in the file so the aligner becomes less likely to derail
\end{itemize}

\chapter{Penn Forced Aligner (Legacy)}\label{penn-forced-aligner-legacy}

\section{Overview}\label{overview-3}

\textbf{What does the Penn Forced Aligner do?} The Penn Forced Aligner takes a sound file and the corresponding transcript of speech and returns a Praat TextGrid with phone and word alignments. You can find the website \href{https://www.ling.upenn.edu/phonetics/old_website_2015/p2fa/}{here}.

\textbf{What does it include?} The system comes with \emph{pre-trained acoustic models} of American English (AE) speech from the SCOTUS corpus, built with the HTK Speech Recognition Toolkit. The section on \href{https://www.eleanorchodroff.com/tutorial/kaldi/introduction.html}{Kaldi} introduces a similar, but alternative system to HTK, in case you'd like to train your own acoustic models from scratch. In addition to the acoustic models, the download also includes a large lexicon of words based on the CMU Pronouncing Dictionary. The dictionary contains over 100,000 words with their standard pronunciation transcriptions in \href{http://en.wikipedia.org/wiki/Arpabet}{Arpabet}. Arpabet is a machine-readable phonetic alphabet of General American English with stress marked on the vowels. While I have, perhaps wrongly, used the Penn Forced Aligner to create phonetic alignments for other languages, those transcriptions had to be forced into Arpabet phones. For a quick alignment, this works fine, but please be advised: the acoustic models of the Penn Forced Aligner are trained on American English, so any speech it is given will be treated like American English. Unless manual adjustments follow the automatic alignment, the PFA alignment would not be ideal for most non-AE phonetic analyses. Again, see the section on \href{https://www.eleanorchodroff.com/tutorial/kaldi/introduction.html}{Kaldi} if you would like to create acoustic models for a different language or dialect.

\section{Prerequisites}\label{prerequisites}

\begin{itemize}
\tightlist
\item
  HTK Toolkit Version 3.4

  \begin{itemize}
  \tightlist
  \item
    Version 3.4.1 will \textbf{not} work
  \item
    Download here: \url{http://htk.eng.cam.ac.uk/}
  \item
    Many run into issues installing this, but detailed installation instructions (and additional tutorial notes) can be found \href{http://linguisticmystic.com/2014/02/12/penn-forced-aligner-on-mac-os-x/}{here}. \textbf{Update} as of November 11, 2018: this website (linguisticmystic.com) no longer works, but covered how to install the Penn Forced Aligner on a Mac. The Penn Forced Aligner is no longer being maintained, and has instead been replaced by FAVE (section \ref{fave-align}). The corresponding prerequisites for HTK installation on Mac are now covered on the \href{https://github.com/JoFrhwld/FAVE/wiki/HTK-on-OS-X}{FAVE wiki}).
  \end{itemize}
\item
  Python 2.5 or 2.6

  \begin{itemize}
  \tightlist
  \item
    Pre-installed on Macs, but you can double-check by opening the terminal and typing python. This will return the version installed. I have been using Version 2.7.6 and have not had any issues running the aligner. Since you've typed python, the terminal is now running with the assumption of python syntax. You want to exit this to get back to its Unix base; to do this, type \texttt{exit()} or \texttt{quit()}. If you have an earlier version of Python, you may need to type \texttt{exit}.
  \end{itemize}
\item
  SoX

  \begin{itemize}
  \tightlist
  \item
    Download here: \url{http://sourceforge.net/projects/sox/}
  \end{itemize}
\end{itemize}

\section{Modifying the lexicon}\label{modifying-the-lexicon}

After downloading the Penn Forced Aligner, you should now have a directory named \texttt{p2fa}. Before beginning any of the following steps, take note of the parent directory and path, as you will need to direct your terminal to the exact \texttt{p2fa} location. For example, my \texttt{p2fa} directory is located in \texttt{/Users/Eleanor}. I can direct my terminal now to that location and view the contents of that directory by typing the following:

\begin{Shaded}
\begin{Highlighting}[]
\NormalTok{cd }\OperatorTok{/}\NormalTok{Users}\OperatorTok{/}\NormalTok{Eleanor}\OperatorTok{/}
\NormalTok{ls}
\end{Highlighting}
\end{Shaded}

Once you've located the \texttt{p2fa} directory, you can find the lexicon in \texttt{p2fa/model/dict}.

Despite the lack of a \texttt{.txt} extension, \texttt{dict} is a text file, making it quite easy to add words and non-words alike. You'll want to make sure that all words in your transcript are indeed in the dictionary. This can be done by converting your transcript into a word list and comparing it against the dictionary.

First, you'll need a copy of your transcript in a text file called \texttt{fulltranscript.txt}. (Actually, you can call it whatever you want; just make sure to change the name in the code!)

In the terminal, navigate to the location of your transcript and then we can use the long code to create a list of all unique words in the transcript. My \texttt{fulltranscript.txt} is located in \texttt{/Users/Eleanor/myCorpus}. You'll need to change that part to match your file location.

\begin{Shaded}
\begin{Highlighting}[]
\NormalTok{cd }\OperatorTok{/}\NormalTok{Users}\OperatorTok{/}\NormalTok{Eleanor}\OperatorTok{/}\NormalTok{myCorpus}
\NormalTok{tr }\StringTok{' '} \StringTok{'}\CharTok{\textbackslash{}n}\StringTok{'} \OperatorTok{<}\StringTok{ }\NormalTok{fulltranscript.txt }\OperatorTok{|}\StringTok{ }\NormalTok{tr }\StringTok{'[a-z]'} \StringTok{'[A-Z]'} \OperatorTok{|}\StringTok{ }\NormalTok{\textbackslash{}}
\NormalTok{sed }\StringTok{'/^$/d'} \OperatorTok{|}\StringTok{ }\NormalTok{sed }\StringTok{'/[.,?!;:]/d'} \OperatorTok{|}\StringTok{ }\NormalTok{sort }\OperatorTok{|}\StringTok{ }\NormalTok{uniq }\OperatorTok{-}\NormalTok{c }\OperatorTok{|}\StringTok{ }\NormalTok{sed }\StringTok{'s/^ *//'} \OperatorTok{|}\StringTok{ }\NormalTok{\textbackslash{}}
\NormalTok{sort }\OperatorTok{-}\NormalTok{r }\OperatorTok{-}\NormalTok{n }\OperatorTok{>}\StringTok{ }\NormalTok{fulltranscript_words.txt}
\end{Highlighting}
\end{Shaded}

The above clearly does a whole slew of functions. It will first take your transcript, separate each word with a new line (\texttt{tr\ \textquotesingle{}\ \textquotesingle{}\ \textquotesingle{}\textbackslash{}n\textquotesingle{}\ \textless{}\ fulltranscript.txt}), capitalize all letters (\texttt{tr\ \textquotesingle{}{[}a-z{]}\textquotesingle{}\ \textquotesingle{}{[}A-Z{]}\textquotesingle{}}), delete blank lines (\texttt{sed\ \textquotesingle{}/\^{}\$/d\textquotesingle{}}), remove punctuation (\texttt{sed\ \textquotesingle{}/{[}.,?!;:{]}/d\textquotesingle{}}), sort the words (\texttt{sort}), remove duplicate words (\texttt{uniq\ -c}), delete blank lines/trailing white space again (\texttt{sed\ \textquotesingle{}s/\^{}\ *//\textquotesingle{}}), sort again and give you a count of how many times each word appears in the transcript (\texttt{sort\ -r\ -n\ \textgreater{}\ fulltranscript\_words.txt}).

The following code will find the word pronunciations in the CMU dictionary. This is accomplished by taking the words and putting them into the regular expression format for locating the beginning of a line (\texttt{\^{}}). This results in \texttt{tmp.txt}. That file is then compared against the CMU dictionary. If the word is in the dictionary, then the dictionary line is extracted such that you have both the word and its pronunciation. Note that when using the \texttt{cut} command, the default cut is tab (\texttt{cut\ -f\ 2}), but if the delimiter is anything other than tab (space, comma, etc.), it can be specified with \texttt{cut\ -d\ \textquotesingle{}my\ delimiter\textquotesingle{}\ -f\ 2\ myfile.txt}.

\begin{Shaded}
\begin{Highlighting}[]
\NormalTok{cut }\OperatorTok{-}\NormalTok{d }\StringTok{' '} \OperatorTok{-}\NormalTok{f }\DecValTok{2}\NormalTok{ fulltranscript_}\OperatorTok{**}\NormalTok{words}\OperatorTok{**}\NormalTok{.txt }\OperatorTok{|}\StringTok{ }\NormalTok{sed }\StringTok{'s/^/^/'} \OperatorTok{|}\StringTok{ }\NormalTok{sed }\StringTok{'s/$/  /'} \OperatorTok{>}\StringTok{ }\NormalTok{tmp.txt}
\NormalTok{egrep }\OperatorTok{--}\NormalTok{file=tmp.txt }\OperatorTok{/}\NormalTok{Users}\OperatorTok{/}\NormalTok{Eleanor}\OperatorTok{/}\NormalTok{p2fa}\OperatorTok{/}\NormalTok{model}\OperatorTok{/}\NormalTok{dict }\OperatorTok{>}\StringTok{ }\NormalTok{fulltranscript_words_pron.txt;}
\end{Highlighting}
\end{Shaded}

You then need to determine which words were skipped in this process, i.e., the words missing from the CMU dictionary. This can be done by comparing the final \texttt{fulltranscript\_words\_pron.txt} against the original \texttt{fulltranscript\_words.txt}.

\begin{Shaded}
\begin{Highlighting}[]
\CommentTok{# extracts and sorts relevant columns and stores in tmp file}
\NormalTok{sort }\OperatorTok{-}\NormalTok{k }\DecValTok{2}\NormalTok{ fulltranscript_words.txt }\OperatorTok{>}\StringTok{ }\NormalTok{tmp1.txt}
\NormalTok{sort }\OperatorTok{-}\NormalTok{k }\DecValTok{1}\NormalTok{ fulltranscript_words_pron.txt }\OperatorTok{>}\StringTok{ }\NormalTok{tmp2.txt}
                    
\CommentTok{# merges the two files to list words, word count, and pron}
\NormalTok{join }\OperatorTok{-}\DecValTok{1} \DecValTok{2} \OperatorTok{-}\DecValTok{2} \DecValTok{1}\NormalTok{ tmp1.txt tmp2.txt }\OperatorTok{>}\StringTok{ }\NormalTok{fulltranscript_words_pron2.txt}
                    
\CommentTok{# lists words with missing pronunciations}
\CommentTok{# these are the words you need to add to the dictionary}
\NormalTok{join }\OperatorTok{-}\NormalTok{v }\DecValTok{1} \OperatorTok{-}\DecValTok{1} \DecValTok{2} \OperatorTok{-}\DecValTok{2} \DecValTok{1}\NormalTok{ tmp1.txt tmp2.txt }\OperatorTok{>}\StringTok{ }\NormalTok{fulltranscript_words_missing_pron.txt}
\end{Highlighting}
\end{Shaded}

\begin{Shaded}
\begin{Highlighting}[]
\CommentTok{# Extras:}
\CommentTok{# get count of word types}
\NormalTok{wc }\OperatorTok{-}\NormalTok{l }\OperatorTok{<}\StringTok{ }\NormalTok{fulltranscript_words.txt}
                        
\CommentTok{# get count of phone types}
\NormalTok{wc }\OperatorTok{-}\NormalTok{l }\OperatorTok{<}\StringTok{ }\NormalTok{fulltranscript_words_pron.txt}
                        
\CommentTok{# get count of phone tokens}
\NormalTok{cut }\OperatorTok{-}\NormalTok{d}\StringTok{' '} \OperatorTok{-}\NormalTok{f }\DecValTok{3}\OperatorTok{-}\StringTok{ }\ErrorTok{<}\StringTok{ }\NormalTok{fulltranscript_words_pron.txt }\OperatorTok{|}\StringTok{ }\NormalTok{tr }\StringTok{' '} \StringTok{'}\CharTok{\textbackslash{}n}\StringTok{'} \OperatorTok{|}\StringTok{ }\NormalTok{sort }\OperatorTok{|}\StringTok{ }\NormalTok{uniq }\OperatorTok{-}\NormalTok{c }\OperatorTok{|}\StringTok{ }\NormalTok{\textbackslash{}}
\NormalTok{sed }\StringTok{'s/^ *//'} \OperatorTok{|}\StringTok{ }\NormalTok{sort }\OperatorTok{-}\NormalTok{r }\OperatorTok{-}\NormalTok{n }\OperatorTok{>}\StringTok{ }\NormalTok{fulltranscript_phones.txt}
\end{Highlighting}
\end{Shaded}

Lexical entries need to be added in the same format as the rest of the dictionary, which is the word in all caps followed by two spaces and the CMU pronunciation with stress on the vowel. For example:

\hypertarget{textfile}{}
KLATT K L AE1 T ELEANOR EH1 L AH0 N AO2 R ELEANOR EH1 L AH0 N ER2

Multiple pronunciation variants are fine (this can even be used to test some interesting hypotheses). CMU provides a nice tool for converting standard spellings of words and non-words into the correct pronunciation. This can be found here: \url{http://www.speech.cs.cmu.edu/tools/lextool.html}.

After running this tool, you will need to add the stress value to the vowels (1 = primary, 0 = unstressed, 2 = secondary). You can always refer to similar words in the dictionary for an example.

\section{Running the aligner}\label{running-the-aligner-2}

The standard implementation of the Penn Forced Aligner will process a single wav file and transcript, returning the phone and word alignment on a Praat TextGrid.

Before beginning this part of the tutorial, make sure that all the words in your transcript are in the included lexicon, \texttt{p2fa/model/dict}. If you're not sure or know that they are not, please visit Section \ref{modifying-the-lexicon}.

Ingredients:\\
* Wav file of recording\\
* Corresponding transcript. Example below:\\

\hypertarget{textfile}{}
SAY TUTT AGAIN

SAY PAT AGAIN

SAY DOT AGAIN

The transcript should contain the words with spaces between them; these are standardly capitalized, but the aligner will accept lowercase and uppercase letters. Line breaks are fine, as are apostrophes, as long as that spelling is in the dictionary. All punctuation should be removed. Others have suggested adding an ``\texttt{sp}'' or space between words and sentences. This is not necessary. The aligner will determiner whether or not a small space or silence is present.

An important note is that the aligner can derail if there is untranscribed noise or speech in the recording. In my experience, it's not too bad at recovering after a short while, but it's best to avoid that situation. This can be accomplished by giving it smaller portions of the wav file, or ensuring that all noise and extraneous speech is explicitly transcribed. Smaller portions of the wav file can be created manually by extracting the relevant clip(s). Alternatively, a modified version of the script can process relevant sections of speech defined by their start and end times. Yet another option is to transcribe noise as \texttt{\{NS\}} and silence as \texttt{\{SP\}}. I have not tried this method, so I do not know how robust it is to extraneous speech.

To process a single wav file and transcript, direct the terminal to the directory containing the Penn Forced Aligner \texttt{align.py} script with \texttt{cd}, then type the second command. The arguments to the align.py script are the locations of the wav file and transcript file. The script creates the aligned TextGrid as output (\texttt{subj01.TextGrid}, but you can call it whatever you want).

\begin{Shaded}
\begin{Highlighting}[]
\NormalTok{cd }\OperatorTok{~}\ErrorTok{/}\NormalTok{p2fa}
\NormalTok{python align.py }\OperatorTok{/}\NormalTok{Users}\OperatorTok{/}\NormalTok{Eleanor}\OperatorTok{/}\NormalTok{myCorpus}\OperatorTok{/}\NormalTok{subj01.wav \textbackslash{}}
\OperatorTok{/}\NormalTok{Users}\OperatorTok{/}\NormalTok{Eleanor}\OperatorTok{/}\NormalTok{myCorpus}\OperatorTok{/}\NormalTok{subj01.txt subj01.TextGrid}
\end{Highlighting}
\end{Shaded}

And that's it!

\chapter{AutoVOT}\label{autovot}

\section{Overview}\label{overview-4}

\textbf{What does AutoVOT do?} AutoVOT takes a sound file and Praat TextGrid marked with the locations of word-initial, prevocalic stop consonants and creates a new tier with boundaries marking the stop consonant burst release and following vocalic onset. With these boundaries, positive VOT can be measured efficiently using a standard acoustic analysis program such as Praat.

What does it include? AutoVOT can be used via Praat directly or via the command line. The system comes with pre-trained acoustic models/classifiers for American English and British English word-initial, prevocalic VOTs. It also offers the option to train your own acoustic models from labeled training data.

As the online tutorial is extremely helpful, I will just offer a few tips from experience. You will first need to visit the online tutorial for the prerequisites and basic installation.

\textbf{Online tutorial and download:} \url{https://github.com/mlml/autovot/}

While not mentioned in the online tutorial, another prerequisite is NumPy. This can be downloaded here: \href{http://wwww.numpy.org}{http://www.numpy.org/}

*Note that this may have been updated since I last used it. The developer of AutoVOT had mentioned he may remove this dependency.

The recipe assumes that you have TextGrids containing the phone and word level transcriptions (e.g., Penn Forced Aligner output)! To run AutoVOT, the word-initial, prevocalic stop consonants need to be located and given a ``window of analysis'' in the TextGrid. The window of analysis is the interval surrounding the stop consonant that AutoVOT will process. The tutorial will cover one method for accomplishing this.

\section{Recipe}\label{recipe}

\begin{enumerate}
\def\labelenumi{\arabic{enumi}.}
\tightlist
\item
  Create a list of word-initial, prevocalic (CV) words in your transcript
\end{enumerate}

There are many ways to accomplish this; the following is just one suggested way using some Unix and the P2FA dictionary. The following bash commands take multiple transcripts with the naming structure \texttt{S001.txt}, \texttt{S002.txt}, etc. as input, concatenates them, removes punctuation, puts each word on a new line, strips end of line characters, converts lowercase to uppercase, then removes duplicate words. The output of this is returned in the text file \texttt{fulltranscript\_words.txt}. If you have a single transcript, you can replace \texttt{S*{[}0-9{]}.txt} with the name of your transcript. Before running this code, you must direct the terminal (\texttt{cd}) to the directory housing your full transcript (\texttt{fulltranscript.txt}).

\begin{Shaded}
\begin{Highlighting}[]
\NormalTok{cat S}\OperatorTok{*}\NormalTok{[}\DecValTok{0}\OperatorTok{-}\DecValTok{9}\NormalTok{].txt }\OperatorTok{|}\StringTok{ }\NormalTok{tr }\OperatorTok{-}\NormalTok{d }\StringTok{'[:punct:]'} \OperatorTok{|}\StringTok{ }\NormalTok{tr }\StringTok{' '} \StringTok{'}\CharTok{\textbackslash{}n}\StringTok{'} \OperatorTok{|}\StringTok{  }
\NormalTok{sed }\StringTok{'/^$/d'} \OperatorTok{|}\StringTok{ }\NormalTok{tr }\StringTok{'[a-z]'} \StringTok{'[A-Z]'} \OperatorTok{|}\StringTok{ }\NormalTok{sort }\OperatorTok{|}\StringTok{ }\NormalTok{uniq }\OperatorTok{>}\StringTok{ }\NormalTok{transcript.txt}
\end{Highlighting}
\end{Shaded}

\href{https://www.eleanorchodroff.com/tutorial/scripts/matchText.py}{matchText.py}

\texttt{matchText.py} takes as input the CMU Pronouncing Dictionary with a `.txt' extension. It identifies which of the words in your transcript begin with stop consonants in prevocalic position. These words are stored in the text file \texttt{wordList.txt}. You will need to modify the path locations and possibly the regular expression in \texttt{matchText.py}.

\begin{Shaded}
\begin{Highlighting}[]
\CommentTok{# run matchText.py}
\NormalTok{python matchText.py}
\end{Highlighting}
\end{Shaded}

\begin{enumerate}
\def\labelenumi{\arabic{enumi}.}
\setcounter{enumi}{1}
\tightlist
\item
  Find start and end times for words on \texttt{wordList.txt} in TextGrids with phone- and word- level boundaries
\end{enumerate}

{[}findWords.praat{]}``(\url{https://www.eleanorchodroff.com/tutorial/scripts/findWords.praat})\{target=''\_blank``\} \texttt{findWords.praat} takes as input \texttt{wordList.txt} and returns the start and end times of matching words in the audio file. This is stored in the text file \texttt{CVWordLocations.txt}. You will need to modify the source and destination paths in \texttt{findWords.praat}. Verify that the regular expression in \texttt{‘Create\ Strings\ as\ file\ list\ldots{}’} will work for your setup.

\begin{enumerate}
\def\labelenumi{\arabic{enumi}.}
\setcounter{enumi}{2}
\tightlist
\item
  Create AutoVOT intervals on a new tier in your Praat TextGrid
\end{enumerate}

\href{https://www.eleanorchodroff.com/tutorial/scripts/makeAutoVOTTextGrids.praat}{makeAutoVOTTextGrids.praat}

This script takes as input the text file \texttt{CVWordLocations.txt} and the Penn Forced Aligner TextGrids. It adds an interval tier `vot' and renames the TextGrids to \texttt{filename\_allauto.TextGrid}. Because all stop consonants are word-initial, the start of the word is assumed to be the start of the stop consonant. The end of the stop consonant is identified using the interval on the phone tier that aligns with the start of the word. Those boundaries are then used to create an AutoVOT check interval on the new vot tier. If the stop consonant begins with PTK, then those boundaries are extended 31 ms in both directions. If the stop consonant begins with BDG, then the boundaries are extended 11 ms in both directions. The extra, odd millisecond is to reduce the chance of placing a boundary where one already exists. Note that the P2FA boundaries can only be placed at 10 ms intervals, so adding time at a factor of 10 ms results in overlapping/identical boundaries. The interval text is then set with the phone name (PTKBDG) after all boundaries have been created. This ensures that overlapping intervals will not be a problem. The output should look something like this:

\begin{figure}
\centering
\includegraphics{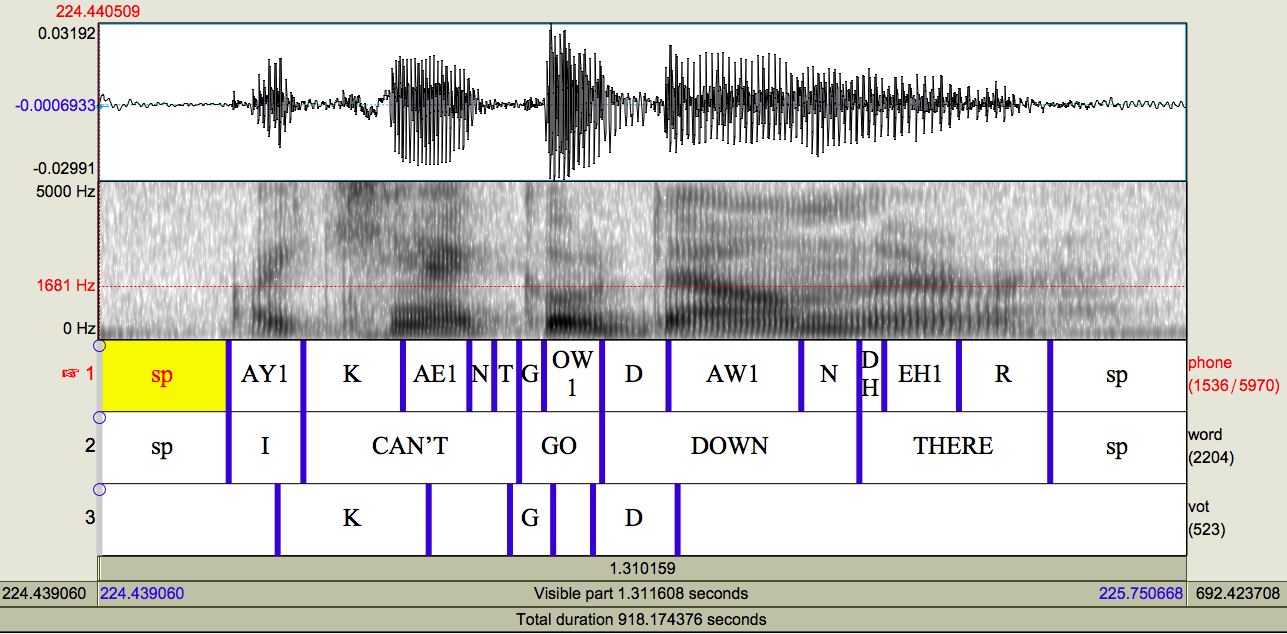}
\caption{Example TextGrid for AutoVOT}
\end{figure}

The different labels in the `vot' tier will have some consequences when running AutoVOT. The AutoVOT analysis depends on a consistent label for the intervals it needs to check. Since there are six different interval labels {[}PTKBDG{]}, AutoVOT will need to be run six different times. For me, this is preferable, as I then know exactly which stop consonant was measured. For some, however, this may be too tedious. In that case, I still highly recommend running the voiced and voiceless stop consonants separately.

\begin{enumerate}
\def\labelenumi{\arabic{enumi}.}
\setcounter{enumi}{3}
\tightlist
\item
  Make a list of your TextGrids and wav files and move them the lists to \texttt{autovot/experiments/config}
\end{enumerate}

The code below will generate a textfile located in the correct autovot folder that contains a list of the TextGrids with their path (\texttt{PWD} is the command that prints this). When generating these lists, the terminal must be in the directory that contains the TextGrids and wav files, respectively.

\begin{Shaded}
\begin{Highlighting}[]
\NormalTok{cd experiment}\OperatorTok{/}\NormalTok{myTextGrids  }
\NormalTok{ls }\OperatorTok{-}\NormalTok{d }\OperatorTok{-}\DecValTok{1} \OperatorTok{$}\NormalTok{PWD}\OperatorTok{/}\ErrorTok{*}\NormalTok{_vot.TextGrid }\OperatorTok{>}\StringTok{ }\ErrorTok{~/}\NormalTok{autovot}\OperatorTok{/}\NormalTok{experiments}\OperatorTok{/}\NormalTok{config}\OperatorTok{/}\NormalTok{ListTextGrids.txt}
                    
\NormalTok{cd experiment}\OperatorTok{/}\NormalTok{myWavFiles}
\NormalTok{ls }\OperatorTok{-}\NormalTok{d }\OperatorTok{-}\DecValTok{1} \OperatorTok{$}\NormalTok{PWD}\OperatorTok{/}\ErrorTok{*}\NormalTok{.wav }\OperatorTok{>}\StringTok{ }\ErrorTok{~/}\NormalTok{autovot}\OperatorTok{/}\NormalTok{experiments}\OperatorTok{/}\NormalTok{config}\OperatorTok{/}\NormalTok{ListWavFiles.txt}
\end{Highlighting}
\end{Shaded}

\begin{enumerate}
\def\labelenumi{\arabic{enumi}.}
\setcounter{enumi}{4}
\tightlist
\item
  Run AutoVOT from the terminal on \textbf{one} stop consonant
\end{enumerate}

Modify the arguments to \texttt{auto\_vot\_decode.py}: \texttt{window\_mark} should be set to the stop consonant to be analyzed. We recommend setting \texttt{—min\_vot\_length} to 4 (ms) for voiced stops and 15 (ms) for voiceless stops. After each stop consonant, there is a post-processing step. Make sure to do that to avoid overwriting the AutoVOT output (see step 6)!

Example code for a voiceless stop:

\begin{Shaded}
\begin{Highlighting}[]
\NormalTok{cd experiments            }
\NormalTok{export PATH=}\ErrorTok{$}\NormalTok{PATH}\OperatorTok{:}\ErrorTok{/}\NormalTok{Users}\OperatorTok{/}\NormalTok{Eleanor}\OperatorTok{/}\NormalTok{autovot}\OperatorTok{/}\NormalTok{autovot}\OperatorTok{/}\NormalTok{bin}
            
\NormalTok{auto_vot_decode.py }\OperatorTok{--}\NormalTok{window_tier vot }\OperatorTok{--}\NormalTok{window_mark P }\OperatorTok{--}\NormalTok{min_vot_length }\DecValTok{15}\NormalTok{ \textbackslash{}}
\NormalTok{config}\OperatorTok{/}\NormalTok{ListWavFiles.txt config}\OperatorTok{/}\NormalTok{ListTextGrids.txt \textbackslash{}}
\OperatorTok{/}\NormalTok{Users}\OperatorTok{/}\NormalTok{Eleanor}\OperatorTok{/}\NormalTok{autovot}\OperatorTok{/}\NormalTok{autovot}\OperatorTok{/}\NormalTok{bin}\OperatorTok{/}\NormalTok{models}\OperatorTok{/}\NormalTok{vot_predictor.amanda.max_num_instances_}\FloatTok{1000.}\NormalTok{model}
\end{Highlighting}
\end{Shaded}

Example code for a voiced stop:

\begin{Shaded}
\begin{Highlighting}[]
\NormalTok{cd experiments}
\NormalTok{export PATH=}\ErrorTok{$}\NormalTok{PATH}\OperatorTok{:}\ErrorTok{/}\NormalTok{Users}\OperatorTok{/}\NormalTok{Eleanor}\OperatorTok{/}\NormalTok{autovot}\OperatorTok{/}\NormalTok{autovot}\OperatorTok{/}\NormalTok{bin}

\NormalTok{auto_vot_decode.py }\OperatorTok{--}\NormalTok{window_tier vot }\OperatorTok{--}\NormalTok{window_mark B }\OperatorTok{--}\NormalTok{min_vot_length }\DecValTok{4}\NormalTok{ \textbackslash{}}
\NormalTok{config}\OperatorTok{/}\NormalTok{ListWavFiles.txt config}\OperatorTok{/}\NormalTok{ListTextGrids.txt \textbackslash{} }
\OperatorTok{/}\NormalTok{Users}\OperatorTok{/}\NormalTok{Eleanor}\OperatorTok{/}\NormalTok{autovot}\OperatorTok{/}\NormalTok{autovot}\OperatorTok{/}\NormalTok{bin}\OperatorTok{/}\NormalTok{models}\OperatorTok{/}\NormalTok{vot_predictor.amanda.max_num_instances_}\FloatTok{1000.}\NormalTok{model}
\end{Highlighting}
\end{Shaded}

The path (\texttt{export\ PATH=\$PATH} command) should always be set from the \texttt{experiments} directory before running AutoVOT. I recommend the above argument specification, but the structure can also be modified to suit your particular dataset. The arguments in the AutoVOT decode command are listed here:

\begin{itemize}
\tightlist
\item
  --window\_tier

  \begin{itemize}
  \tightlist
  \item
    This refers to the TextGrid tier that contains the intervals to check (or windows of analysis). The current procedure has called this tier `vot'
  \end{itemize}
\item
  --window\_mark

  \begin{itemize}
  \tightlist
  \item
    This refers to the label of the interval to check. The current procedure has six different labels {[}PTKBDG{]}, so this command will need to be run six times, once for each of these labels. After each run, the output tier will need to be renamed so that it is not overwritten.
  \end{itemize}
\item
  --min\_vot\_length

  \begin{itemize}
  \tightlist
  \item
    This refers to the minimum allowed VOT length. I would recommend 15ms for voiceless stops and 4ms for voiced stops, but this can be modified. It should be noted, however, that performance degrades on the voiceless stop measurements if the minimum VOT is too low. (This is why I recommend running AutoVOT separately for the voiced and voiceless stops.)
  \end{itemize}
\item
  Path from experiments to the list of wav files
\item
  Path from experiments to the list of TextGrids
\item
  Path to AutoVOT classifier

  \begin{itemize}
  \tightlist
  \item
    The default classifier is the one named \texttt{amanda}, but there are a few others you can try. While the \texttt{amanda} classifier is hidden to the user, the others are located in the \texttt{autovot/bin/models} folder. Alternatively, AutoVOT gives you the option to train your own classifier on labeled data. For more information on training, please visit their main website: \url{https://github.com/mlml/autovot/}.
  \end{itemize}
\end{itemize}

\begin{enumerate}
\def\labelenumi{\arabic{enumi}.}
\setcounter{enumi}{5}
\tightlist
\item
  Rename AutoVOT output tier
\end{enumerate}

\href{https://www.eleanorchodroff.com/tutorial/scripts/autoVOTpostproc.praat}{autoVOTpostproc.praat}

After each stop consonant is processed, run \texttt{autoVOTpostproc.praat} to rename the AutoVOT output tier. Otherwise, AutoVOT will overwrite your previous work.

Make sure to change the phone name in the script.

Return to step 5 and repeat until all 6 stop consonants have been processed.You will need to modify the path to the TextGrids, the tier labels, and the new tier name. Once you have completed this, the cycle starts over until you have all six stops.

\begin{enumerate}
\def\labelenumi{\arabic{enumi}.}
\setcounter{enumi}{6}
\tightlist
\item
  Move all AutoVOT boundaries to one tier
\end{enumerate}

\href{https://www.eleanorchodroff.com/tutorial/scripts/resetBoundaries_stops.praat}{resetBoundaries\_stops.praat}

After running AutoVOT, you should now have a TextGrid with 6 different output tiers: one for each stop consonant. These tiers can be collapsed into one tier with resetBoundaries\_stops.praat. You will need to modify the path directory. In addition, if your AutoVOT output does not occupy tiers 3-8, you will need to modify the tier numbers in the script.

The final product should look like this:

\begin{figure}
\centering
\includegraphics{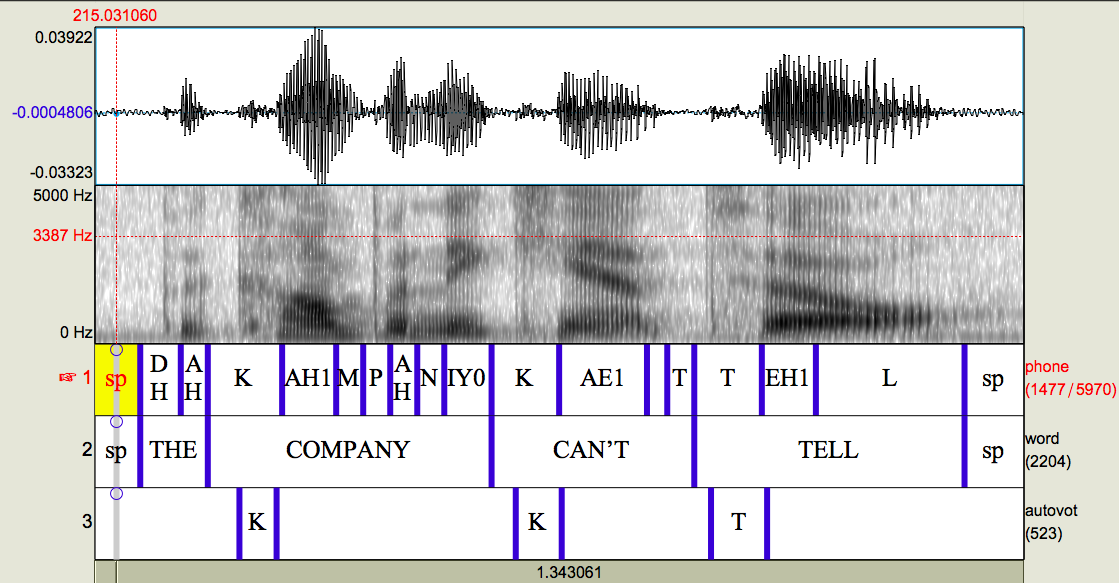}
\caption{Example output TextGrid from AutoVOT}
\end{figure}

This resembles the previous picture of the TextGrid, but note that the boundaries on the autovot tier are now located at the burst and vocalic onset in the signal.

****If you have manually placed/corrected boundaries, continue on. Otherwise, skip to step 11!****

\begin{enumerate}
\def\labelenumi{\arabic{enumi}.}
\setcounter{enumi}{7}
\tightlist
\item
  Stack TextGrids with manual boundaries and the AutoVOT boundaries
\end{enumerate}

\href{https://www.eleanorchodroff.com/tutorial/scripts/stackTextGrids.praat}{stackTextGrids.praat}

This script takes as input TextGrids with manual boundaries (we have two different types) and the AutoVOT TextGrids (\texttt{\_stops.TextGrid}). It places all the tiers together in one TextGrid and renames the file \texttt{\_stacked.TextGrid}.

We have two different TextGrids with manual boundaries in them: \texttt{\_autovot.TextGrid} and \texttt{\_check.TextGrid}. Using the \texttt{\_autovot} TextGrids was meant to eliminate bias from seeing the AutoVOT output. Those TextGrids only contain the window of analysis and not the final measurement. On the other hand, manual adjustments on the \texttt{\_check} TextGrids were made directly to the AutoVOT output. This was for efficiency. Only the boundaries on the \_autovot files were used for comparison to the AutoVOT output.

****If you want to compare AutoVOT and manual measurements, complete step 9; otherwise, continue on to step 10.****

\begin{enumerate}
\def\labelenumi{\arabic{enumi}.}
\setcounter{enumi}{8}
\tightlist
\item
  Compare manual and automatic boundaries
\end{enumerate}

\href{https://www.eleanorchodroff.com/tutorial/scripts/measureVOT.praat}{measureVOT.praat}

This script takes as input the \texttt{\_stacked.TextGrid} and creates the file \texttt{manualVOTs.txt}.

\begin{enumerate}
\def\labelenumi{\arabic{enumi}.}
\setcounter{enumi}{9}
\tightlist
\item
  Replace automatic boundaries with manual ones where available
\end{enumerate}

\href{https://www.eleanorchodroff.com/tutorial/scripts/resetBoundaries_stacked.praat}{resetBoundaries\_stacked.praat}

This script takes as input the \texttt{\_stacked.TextGrids} and creates as output \texttt{\_stacked2.TextGrid}.

\begin{enumerate}
\def\labelenumi{\arabic{enumi}.}
\setcounter{enumi}{10}
\tightlist
\item
  Measure VOT and sentence rate
\end{enumerate}

\href{https://www.eleanorchodroff.com/tutorial/scripts/cueAnalysis_new.praat}{cueAnalysis\_new.praat}

This script measures the duration of each burst (positive VOT, if you will), following vowel, and word. It also measures the speaking rate, defined as the average word duration per sentence. The speaking rate component relies on there being two `sp' or silent intervals between each sentence. If your data does not meet this criterion, you can modify the script to fit your data or simply comment out the speaking rate measurement. You will need to modify the path to the wav files and TextGrids.

\chapter{Other resources}\label{other-resources}

There are many other freely available phonetics tools and resources. I've listed just a few of them here. If you know of others that should be added, please let me know!

\begin{itemize}
\tightlist
\item
  \href{https://github.com/MontrealCorpusTools/Montreal-Forced-Aligner}{Montreal Forced Aligner}
\item
  \href{https://github.com/MontrealCorpusTools/speechcorpustools}{Montreal Corpus Tools}
\item
  \href{http://prosodylab.org/tools/aligner/}{Prosodylab Aligner}
\item
  \href{http://www.seas.ucla.edu/spapl/voicesauce/}{VoiceSauce}
\item
  \href{http://www.homepages.ucl.ac.uk/~uclyyix/ProsodyPro/}{ProsodyPro}
\item
  SPAAT
\item
  \href{http://fave.ling.upenn.edu/}{FAVE}
\item
  \href{http://phonologicalcorpustools.github.io/CorpusTools/}{Phonological CorpusTools}
\item
  \href{http://korean.utsc.utoronto.ca/kpa/}{Korean Phonetic Aligner}
\item
  \href{https://www.phonetik.uni-muenchen.de/forschung/Verbmobil/VM14.7eng.html}{Munich AUtomatic Segmentation System (MAUS)}
\item
  \href{https://www.phon.ca/phontrac}{Phon}
\item
  \href{http://phonotactics.anu.edu.au/}{World Phonotactics Database}
\item
  \href{http://www.phonetics.ucla.edu/index/sounds.html}{UCLA Database of Sounds}
\item
  \href{http://mattwinn.com/praat.html}{Praat scripts} by Matt Winn
\item
  \href{http://www.holgermitterer.eu/research.html}{Praat scripts} by Holger Mitterer
\item
  \href{http://savethevowels.org/praat/}{Praat tutorial} by Will Styler
\end{itemize}

\bibliography{book.bib}

\end{document}